\pdfoutput=1

\PassOptionsToPackage{table}{xcolor}
\documentclass[11pt]{article}
\usepackage{ACL2023}

\usepackage{times}
\usepackage{latexsym}
\usepackage{amsmath}
\usepackage{amssymb}
\usepackage{array, multirow, bigdelim, makecell, booktabs} 
\usepackage{cleveref}
\usepackage{graphicx}
\usepackage{subcaption}
\usepackage{adjustbox}

\usepackage{amsmath,amsfonts,bm}

\def\eqref#1{equation~\ref{#1}}

\def\1{\bm{1}}

\def\rr{{\textnormal{r}}}

\def\ry{{\textnormal{y}}}
\def\rz{{\textnormal{z}}}

\DeclareMathAlphabet{\mathsfit}{\encodingdefault}{\sfdefault}{m}{sl}
\SetMathAlphabet{\mathsfit}{bold}{\encodingdefault}{\sfdefault}{bx}{n}

\def\sD{{\mathbb{D}}}

\def\sR{{\mathbb{R}}}

\usepackage[T1]{fontenc}
\usepackage[utf8]{inputenc}
\usepackage{microtype}
\usepackage{inconsolata}

\title{BLIND: Bias Removal With No Demographics}

\author{
    ~~~~~~Hadas Orgad
    \hspace{8em}
    Yonatan Belinkov\thanks{~~Supported by the Viterbi Fellowship in the Center for Computer Engineering at the Technion.} 
    \\
    \texttt{orgad.hadas@cs.technion.ac.il}
    \hspace{2em} 
    \texttt{belinkov@technion.ac.il}
    \vspace{0.3em} \\
    Technion -- Israel Institute of Technology
 }

\begin{document}
\maketitle
\begin{abstract}

Models trained on real-world data tend to imitate and amplify social biases. Common methods to mitigate biases require prior information on the types of biases that should be mitigated (e.g., gender or racial bias) and the social groups associated with each data sample. In this work, we introduce BLIND, a method for bias removal with no prior knowledge of the demographics in the dataset. While training a model on a downstream task, BLIND detects biased samples using an auxiliary model that predicts the main model's success, and down-weights those samples during the training process. Experiments with racial and gender biases in sentiment classification and occupation classification tasks demonstrate that BLIND mitigates social biases without relying on a costly demographic annotation process. Our method is competitive with other methods that require demographic information and sometimes even surpasses them.\footnote{Our code is available at \url{https://github.com/technion-cs-nlp/BLIND}.}

\end{abstract}

\section{Introduction}

Neural natural language processing (NLP) models are known to suffer from social biases, such as performance disparities between genders or races \cite{blodgett-etal-2020-language}. Numerous debiasing methods have been proposed in order to address this issue, with varying degrees of success. A disadvantage of these methods is that they require knowledge of the biases one wishes to mitigate (e.g., gender bias) \cite{bolukbasi2016man, zhao-etal-2018-gender, dearteaga-bios, hall-maudslay-etal-2019-name}. Moreover, some methods require additional annotations for identifying the demographics for each sample in the data, such as the race of the writer \cite{han2021balancing, pmlr-v139-liu21f,pmlr-v162-ravfogel22a, shen-etal-2022-optimising}. Some annotations can be automatically collected from the data, while others require manual annotations or expert knowledge, which can be very costly. Thus, existing methods are typically limited to a small number of datasets and tasks. In this paper, we propose a new debiasing method, \textbf{BLIND} -- \textbf{B}ias remova\textbf{L} w\textbf{I}th \textbf{N}o \textbf{D}emographics.

\begin{figure}[!t]
\begin{subfigure}[h]{\columnwidth}
    \includegraphics[width=1\linewidth]{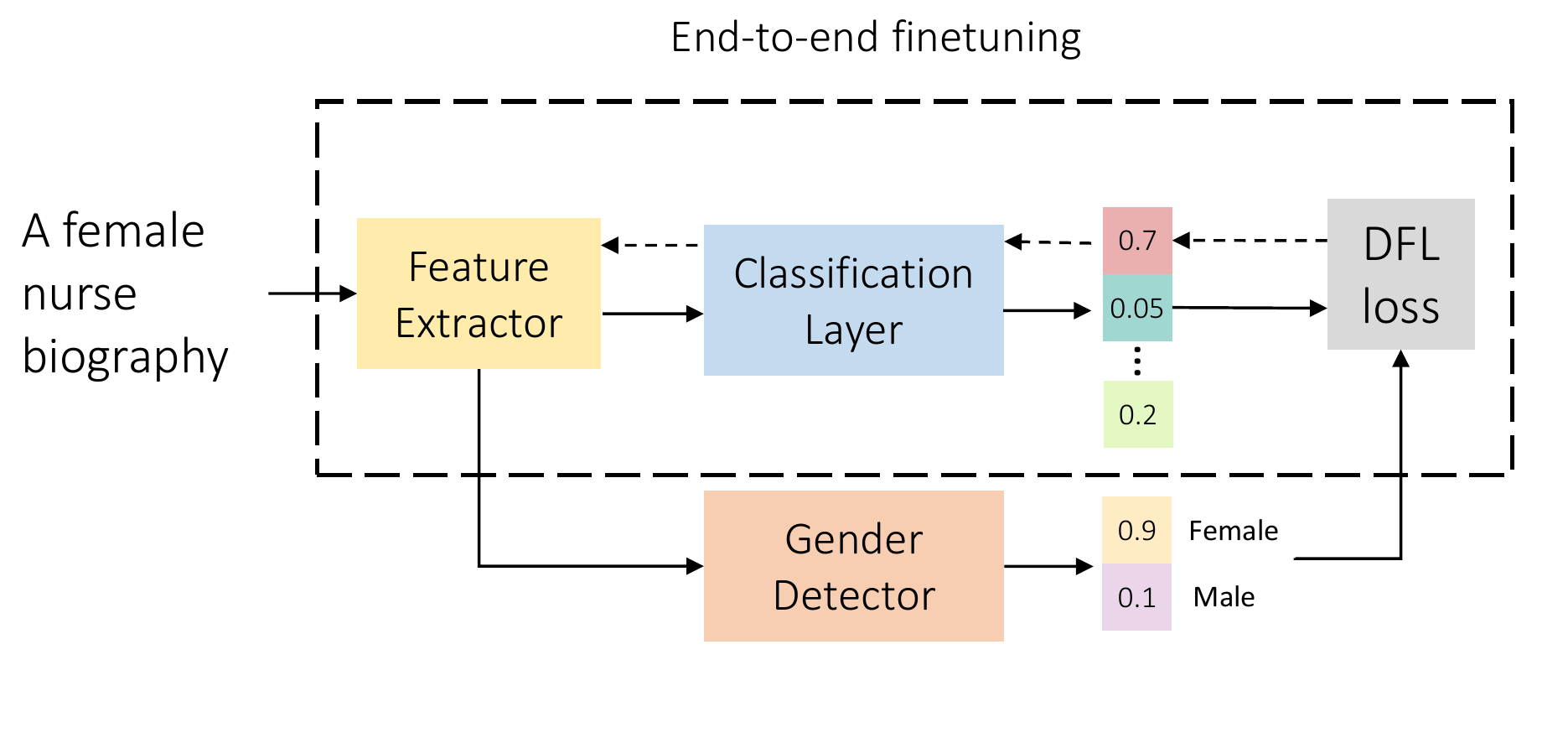}
  \caption{With demographic annotations. Demographics detector learns to predict the demographic data, e.g., gender.}
  \label{fig:framework1}
\end{subfigure}
\begin{subfigure}[h]{\columnwidth}
    \includegraphics[width=1\linewidth]{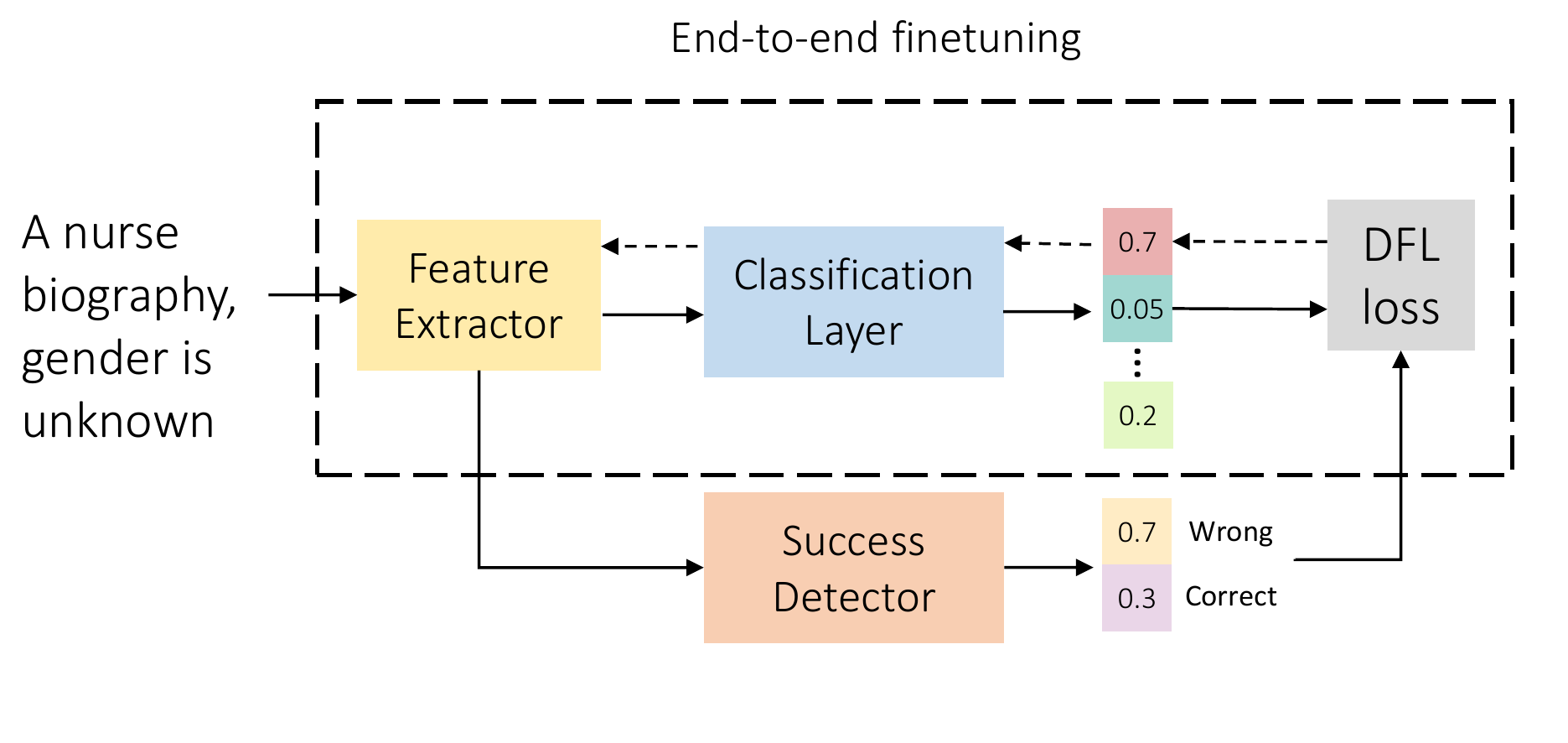}
  \caption{BLIND: Without demographic annotations. Success detector learns to predict when the main model is correct. Supervision is based only on the downstream task labels.}
  \label{fig:framework2}
\end{subfigure}\vspace{1em}
\centering
\begin{subfigure}[h]{.5\columnwidth}
    \includegraphics[width=1\linewidth]{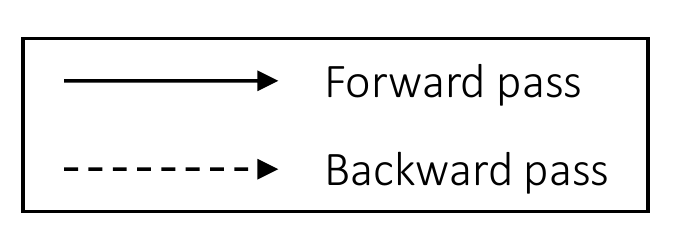}
\end{subfigure}
\caption{Our proposed debiasing methods. In both cases, an auxiliary classifier is trained to detect samples where demographic features may be used as shortcuts and their importance to the main model is down-weighted.}
\end{figure}

We see social bias as a special case of robustness issues resulting from shortcut learning \cite{geirhos2020shortcut}. Our goal is to down-weight samples that contain  demographic features that may be used as shortcuts in downstream tasks.
We first consider a case where we have demographic annotations for every sample in the training set, and train a \emph{demographics detector} --  an auxiliary classifier that takes the main model's representations and predicts the demographic attribute. Then, we down-weight the importance of samples on which the classifier is confident (Figure \ref{fig:framework1}). To our knowledge, this is the first work to consider demographic information for re-weighting samples during training.

When we do not have demographic annotations, 
we make the following observation: The main model has an easier job, or otherwise makes predictable mistakes, when demographic features are used as shortcut features. Thus, we train a \emph{success detector} -- another auxiliary classifier, which takes the representations of the main model and predicts its success on the task. A correct prediction by the success detector means the main model made a shallow decision, since it is possible predict its success or failure without access to the main model's task labels. In such cases we expect that the main model relies on simple, shortcut features, and we use the success detector's confidence to down-weight such samples in the training data (Figure \ref{fig:framework2}). We call this method BLIND.

In both cases, we adapt the debiased focal loss  \cite[DFL;][]{karimi-mahabadi-etal-2020-end}, originally proposed for mitigating annotation artifacts, to down-weight samples that the detectors predicted correctly. In contrast to the original DFL work, which explicitly defined biases to mitigate, we design the bias detection mechanism in a general manner such that the model's biases are learned and mitigated dynamically during training.

We perform experiments on two English NLP tasks and two types of social demographics: occupation classification with gender, and sentiment analysis with race. Our methods successfully reduce bias, with BLIND sometimes succeeding in cases where other methods that use demographic information fail. Our analysis shows that BLIND reduces demographic information in the model's internal representation, even though it does not have access to it. Additionally, BLIND is particularly effective at mitigating bias due to its down-weighting of easy training samples, rather than relying on demographic information alone. This suggests that BLIND may be more robust in mitigating bias than other methods.

\section{Methodology}
\label{sec:methodology}

\subsection{Problem Formulation}

We consider general multi-class classification problems. The dataset $\sD = \{x_i, y_i, z_i\}_{i=1}^N$ consists of triples of input $x_i \in \mathcal{X}$, label $y_i \in \mathcal{Y}$, and a protected attribute $z_i \in \mathcal{Z}$, which corresponds to a demographic group, such as gender. $z$ might be latent, meaning that it cannot be accessed during training or validation stages. Our goal is to learn a mapping $f_M: \mathcal{X} \rightarrow \mathbb{R}^{|Y|}$, such that $f_M$, which we call the main model, is robust to differences in demographics as induced by $z_i$. 

The robustness of a model is measured using a variety of fairness metrics. A fairness metric is a mapping from a model's predictions and the protected attributes associated with each sample to a numerical measure of bias: $M: (\mathbb{R^{|\mathcal{Y}|}}, \mathcal{Z}) \rightarrow \mathbb{R}$. The closer the absolute value is to 0, the  fairer the model. The practical fairness metrics measured in this work are described in \Cref{sec:metrics}.

\subsection{Debiased Focal Loss for Social Bias}

Debiased focal loss was proposed by \citet{karimi-mahabadi-etal-2022-prompt} to improve natural language understanding models on out-of-distribution data. The authors explicitly defined the biases they aim to mitigate, and trained an auxiliary model on the same task as the main model by feeding it with biased features. We model biased samples differently: instead of learning the same downstream task as the main model, our auxiliary model learns a separate task that indicates bias.

The model $f_M$ can be written as a composition of two functions: $g$, the text encoder, and $h_M$, the classifier, such that $f_M = h_M \circ g$. In our case, $g$ is a transformer language model such as BERT \cite{devlin2018bert}, and $h_M$ is a linear classification layer.

\paragraph{Loss term.} We use the DFL formulation to re-weight samples that contain bias. To determine the re-weighting coefficients, we need a separate model that acts as a bias detector, $f_B = h_B \circ g$.
The next two sub-sections define two options for the bias detector, with and without using demographics.

The main and auxiliary models have parameters $\theta_M$ and $\theta_B$ respectively, and the parameters of the encoder $g$ are included in $\theta_M$. The training loss is defined as:
\begin{align}
&\mathcal{L}(\theta_M; \theta_B) = \label{eqn:dfl_formulation}\\
&\Bigl( 1 - \sigma\bigl(f^s_B(x;\theta_B)\bigr)  \Bigr)^\gamma log ( \sigma(f_M^y(x;\theta_m)) \nonumber
\label{eq:loss}
\end{align}

for a single instance $(x,y,s)$, where $\sigma(u)={e^{u^j}} / {\sum_{k=1}^{|\mathcal{Y}|}e^{u^k}}$ is the softmax function, and $\gamma$ is a hyper-parameter that controls the strength of the re-weighting.
Here $s$ is either the demographic attribute $z$ when we have it, or a success indicator of the main model on $x$, as explained below. 
When the bias detector assigns a high probability to $s$, the contribution of this sample to the loss is decreased, and this effect is magnified by $\gamma$ ($\gamma = 0$ restores the vanilla cross-entropy loss).
Both models are trained simultaneously, but only the main model's loss is back-propagated to the encoder $g$, avoiding bias propagation from $f_B$.

\subsection{Debiasing With Demographic Annotations}
\label{sec:debiasing_with_demog}

When demographic attributes are available, we define bias as how easily demographic information can be extracted from a sample. This strategy aligns with the observation by \citet{orgad-etal-2022-gender} that this measure correlates with gender bias metrics. The bias detector is thus formulated as $f_B: g(\mathcal{X}) \rightarrow \sR^{|\mathcal{Z}|}$, taking as input the representations from  $g$ and predicting the demographic attribute; In other words, $s:=z$ in the formulation in \Cref{eqn:dfl_formulation}. \Cref{fig:framework1} illustrates this approach. 

By applying this method, samples in which the demographics detector is successful in predicting demographics ($\sigma(f^z_B(x))$ is high) are down-weighted, and difficult samples ($\sigma(f^z_B(x))$ is low) are up-weighted. Intuitively, the main model is encouraged to focus on samples with less demographic information, which reduces the learning of demographics--task correlations.

\paragraph{Connection to adversarial learning.} While the concept of a demographics model may resemble that used in work on debiasing via adversarial training \cite{zhang2018mitigating, elazar-goldberg-2018-adversarial, han-etal-2021-diverse}, our DFL approach is fundamentally different: rather than reversing gradients from the demographics model to remove information from the main model, we use the demographics model to reweight the loss. Further discussion regarding adversarial learning can be found in Appendix \ref{app:adversarial_learning}.

\subsection{Debiasing Without Demographic Annotations}
\label{sec:debiasing_without_demog}

One of the main weaknesses of other debiasing methods in the field, including the method described in \Cref{sec:debiasing_with_demog}, is the requirement to collect demographic annotations $z$ per data point. These annotations may be expensive or impossible to obtain. Additionally, these annotations often do not address nuances, such as intersectional biases to multiple groups (e.g., both gender and race), or non-binary gender. We propose BLIND as a method for reducing biases without demographic annotations. In this setting, the auxiliary model $f_B$ is trained as a \emph{success detector}. The success detector predicts, for each training sample, whether the main model would be successful at predicting the correct label for the main task. The success detector has no knowledge of the original task. It is formulated as $f_B: g(\mathcal{X}) \rightarrow \sR^{|\mathcal{S}|}$, and $s$ is defined as an indicator function: $s := \mathbb{I}_{f_M(x) == y}$. $s$ is dynamic, and changes across different epochs of the training process. BLIND is illustrated in \Cref{fig:framework2}.

We expect that if the success detector can predict whether the main model is correct or incorrect on a sample (i.e., $\sigma(f^s_B(x)))$ is high), without knowing the task at hand, then the sample contains some simple but biased features, and thus should have reduced weight in the loss. This intuition aligns with the claim that in the context of complex language understanding tasks, all simple feature correlations are spurious \cite{gardner-etal-2021-competency}. \footnote{We also investigated a slightly different variation, in which we only penalize the samples where the success detector was accurate and also the main model was accurate, but we found it to be less effective.}

\section{Experiments}
\subsection{Tasks and Models}

We experiment with two English classification tasks and bias types:

\paragraph{Sentiment Analysis and Race.} We follow the setting of \citet{elazar-goldberg-2018-adversarial}, who used a twitter dataset collected by \citet{blodgett-etal-2016-demographic} to study dialectal variation in social media. We use a subset of 100k samples. As a proxy for the writer's racial identity, each tweet is associated with a label about whether it is written in African American English (AAE) or Mainstream US English (MUSE; often called Standard American English, SAE) based on the geo-location of the author. \citet{elazar-goldberg-2018-adversarial} used emojis in the tweets to derive sentiment labels for the classification task.

\paragraph{Occupation Classification and Gender.} We use the dataset by \citet{dearteaga-bios}, a collection of 400K biographies scraped from the internet to study gender bias in occupation classification. The task is predicting one's occupation based on a subset of their biography -- without the first sentence, which states the occupation. The protected attribute is gender, and each instance is assigned binary genders based on the pronouns in the text, which indicate the individual self-identified gender.

\subsection{Metrics}
\label{sec:metrics}

As recommended by \citet{orgad-belinkov-2022-choose}, who showed that different fairness metrics react differently to debiasing methods, we measure multiple metrics to quantify bias in downstream tasks. They can be separated to two main groups:

\subsubsection{Performance gap metrics}
These indicate the difference in performance between two demographic groups, such as females versus males.

\paragraph{Absolute gap.} For example, for gender we measure the True Positive Rate (TPR) gap for label $y$ as $GAP_{TPR,y} = |TPR_y^F - TPR_y^M|$. We also measure the False Positive Rate (FPR) and Precision gaps. On a multi-class setting, we calculate the absolute sum of gaps across the different labels of the task (denoted $\text{TPR}_s$). We also measure the root mean square for TPR gap (denoted $TPR_{RMS}$) of the gaps, $ \sqrt{\frac{1}{|\mathcal{Y}|}\sum_{y \in \mathcal{Y}} (GAP_{TPR, y})^2}$, since it was used in studies of other debiasing methods we compare to \cite{ravfogel-etal-2020-null, pmlr-v162-ravfogel22a}.

\paragraph{Gaps correlation with training statistics.} When feasible, we compute the Pearson correlation between the gap for each class and the training dataset statistics for this class (denoted $\text{TPR}_p$). For example, the pearson correlation $GAP_{TPR,y}$ and the percentage of female instances in class $y$, as appears in the training set.

\subsubsection{Statistical metrics}

Another family of fairness metrics are statistical metrics, which are measured on probability distributions. To describe these metrics, we use the notation from \Cref{sec:methodology}, and denote the model's predictions with $r$.

\paragraph{Independence.} Measures the statistical dependence between the model's prediction and the protected attributes, by measuring the Kullback–Leibler divergence between two distributions: $KL(P(\rr), P(\rr|\rz=z)), \forall z \in \mathcal{Z}$. We sum the values over $z$ to achieve a single number that describes the independence of the model. This metric does not consider the gold labels, and intutively just measures how much the model's behavior is different on different demographics. 

\paragraph{Separation.} Measures the statistical dependence between the model's prediction given the target label and the protected attributes: $KL(P(\rr|\ry=y), P(\rr|\ry=y, \rz=z)), \forall y \in \mathcal{Y}, z \in \mathcal{Z}$. We sum the values over $y$ and $z$ to achieve a single number. This metric is closely related to TPR and FPR gaps, and intuitively measures if the model behaves differently on each class and demographics.

\paragraph{Sufficiency.} Measures the statistical dependence between the target label given the model's prediction and the protected attributes: $KL(P(\ry|\rr=r), P(\ry|\rr=r,\rz=z)), \forall r \in \mathcal{Y}, z \in \mathcal{Z}$. We sum the values over $r$ and $z$ to achieve a single number. This metric is related to calibration in classification and to precision gap, and can intuitively measure if a model over-promotes or penalizes a certain demographic group.

\subsection{Training and Evaluating}

We experiment with BERT \cite{devlin2018bert} and DeBERTa-v1 \cite{he2020deberta} based architectures, where the transformer model is used as a text encoder and its output and is fed into a linear classifier. We fine-tune the text encoder with the linear layer on the downstream task.

During training, we often use a temperature  $t$ in the softmax function of the auxiliary model $f_B$, which we found to improve training stability. For hyper-parameter search, since we are interested in balancing the importance of task accuracy and fairness, we adapt the `distance to optimum' (DTO) criterion introduced by \citet{han2021balancing}. The DTO calculation is explained in \Cref{app:dto}. We perform model selection without requiring a validation set with demographic annotations, by only choosing the most radical hyper-parameters (highest $\gamma$ and lowest $t$), while limiting the reduction in accuracy (see \Cref{app:gamma}). We chose 0.95 of the maximum achieved accuracy on the task as a threshold, per architecture.
 For more details on the training and evaluation process, see \Cref{app:training}.

All of our models are tested on a balanced dataset (via subsampling), i.e., each label contains the same number of samples from each demographic group. This neutralizes bias in the test dataset, allowing us to truly assess bias in the models, as suggested by \citet{orgad-belinkov-2022-choose}.

\subsection{Baselines and Competitive Systems}

We compare the following training methods:
\begin{description}
    \item[\textbf{DFL-demog (ours)}] DFL trained with demographic annotations, as described in \Cref{sec:debiasing_with_demog}.
    \item[\textbf{BLIND (ours)}] DFL trained without demographic annotations, as described in \Cref{sec:debiasing_without_demog}.
    \item[\textbf{Control}] To rule out any potential form of unintended regularization in BLIND, a control model is trained using random labels for the auxiliary model. We expect this method to have no significant debiasing effect.
    \item[\textbf{Finetuned}] The same model architecture, optimized to solve the downstream task without any debiasing mechanism.
    \item[\textbf{INLP} \citep{ravfogel-etal-2020-null}] A post-hoc method to remove linear information from the neural representation, by repeatedly training a linear classifier to predict a property (in our case, gender or race) from the neural representation, and then projecting the neural representations to the null space of the linear classifier.
    \item[\textbf{RLACE} \citep{pmlr-v162-ravfogel22a}] The goal of this method is also to linearly remove information from the neural representations of a trained model by utilizing a different strategy based on a linear minimax game.
    \item[\textbf{Scrubbing} \cite{dearteaga-bios}] A common approach used to remove gender bias in the occupation classification dataset, is to automatically remove any gender indicators from it, such as ``he'', ``she'', ``Mr.'' or ``Mrs.'', and names. We apply this method on the occupation classification task and also experiments with combining it with our methods (marked as \textbf{+Scrubbing}).
    \item [\textbf{FairBatch} \cite{rohfairbatch}] This method adaptively selects minibatch sizes for improving fairness, with three variations, targeting equal opportunity, equalized odds, and demographic parity.\footnote{See \ref{app:metrics} for more information on the metrics.} The method is designed on binary tasks, thus we apply FairBatch to the sentiment classification task. For a fair comparison, we present the variation that achieved the best fairness metrics we measured.
    \item[\textbf{JTT} \cite{pmlr-v139-liu21f}] Just Train Twice (JTT) is a two-stage train-retrain approach that first trains a model and then trains a second model that upweights misclassified training samples. It works without demographic annotations but requires them for model selection. For a fair comparison, we select the model that has the closest accuracy to our BLIND method.\footnote{We break ties by doing model selection on the validation set with demographic attributes.}
 
\end{description}

\section{Results}

\begin{table*}[t]
\centering
\begin{subtable}{\textwidth}
\adjustbox{max width=\textwidth}{%
\begin{tabular}{lllll|llll}
\toprule
{} & \multicolumn{4}{c}{BERT} & \multicolumn{4}{c}{DeBERTa} \\
\cmidrule(lr){2-5} \cmidrule(lr){6-9}
{} &   Acc $\uparrow$ &   $\text{TPR}_{RMS}$ $\downarrow$ &  Indep $\downarrow$ &  Suff $\downarrow$ & Acc $\uparrow$ &   $\text{TPR}_{RMS}$ $\downarrow$ &  Indep $\downarrow$ &  Suff $\downarrow$ \\
\midrule
Finetuned    & 0.779 & 0.267 & 0.045 & 0.027 & 0.775 & 0.270 & 0.047 & 0.032 \\
INLP $\dagger$         & 0.756* & 0.196* & 0.022* & 0.014* & 0.633* & 0.086* & 0.010* & 0.015* \\
RLACE $\dagger$        & 0.735* & 0.142* & 0.010* & 0.008* & 0.748* & 0.168* & 0.016* & 0.011* \\
FairBatch $\dagger$ &  0.761* & 0.210* & 0.019* & 0.016* & 0.763* & 0.203* & 0.016* & 0.017* \\
$JTT^+$ & 0.767* & 0.194* & 0.016* & 0.016* & 0.780 & 0.191* & 0.013* & 0.014* \\
\midrule
DFL-demog $\dagger$          & 0.762* & 0.124* & 0.000* & 0.003* & 0.766 & 0.177* & 0.009* & 0.008* \\
BLIND & 0.761* & 0.240* & 0.030* & 0.021* & 0.782 & 0.243* & 0.028* & 0.022* \\
\midrule 
Control & 0.776 & 0.480 & 0.046 & 0.026 & 0.780 & 0.253 & 0.038 & 0.025  \\
\bottomrule
\end{tabular}
}
\caption{Sentiment classification.}
\label{tbl:moji}
\end{subtable}

\medskip

\centering
\begin{subtable}{\textwidth}
\adjustbox{max width=\textwidth}{%
\begin{tabular}{lllll|llll}
\toprule
{} & \multicolumn{4}{c}{BERT} & \multicolumn{4}{c}{DeBERTa} \\
\cmidrule(lr){2-5} \cmidrule(lr){6-9}
{} &   Acc $\uparrow$ &   $\text{TPR}_{RMS}$ $\downarrow$ &  $\text{TPR}_p$ $\downarrow$ &  Suff $\downarrow$ & Acc $\uparrow$ &   $\text{TPR}_{RMS}$ $\downarrow$ &  $\text{TPR}_p$ $\downarrow$ &  Suff $\downarrow$ \\
\midrule
Finetuned & 0.864 & 0.136 & 0.809 & 1.559         & 0.864 & 0.134 & 0.819 & 1.597 \\
INLP $\dagger$ & 0.853* & 0.131 & 0.782 & 1.216   & 0.852* & 0.122 & 0.730* & 1.050* \\
RLACE $\dagger$ & 0.866 & 0.131 & 0.809 & 1.413   & 0.868* & 0.126 & 0.808 & 1.361* \\
Scrubbing $\dagger$ & 0.863 & 0.123* & 0.704* & 0.901* & 0.858* & 0.107* & 0.729* & 0.788* \\
$JTT^+$ & 0.846* & 0.136 & 0.761* & 1.417 & 0.841* & 0.140 & 0.770* & 1.398 \\
\midrule 
DFL-demog $\dagger$ & 0.834* & 0.122* & 0.602* & 0.709* & 0.829* & 0.113* & 0.619* & 0.794* \\
+ Scrubbing & 0.830** & 0.111** & 0.569** & 0.584** & 0.841** & 0.114* & 0.578** & 0.592** \\
BLIND & 0.826* & 0.137 & 0.694* & 1.097*   & 0.835* & 0.123 & 0.638* & 0.906* \\
+ Scrubbing & 0.844** & 0.132* & 0.664* & 1.070* & 0.843** & 0.116* & 0.648* & 0.712* \\
\midrule 
Control & 0.864 & 0.140 & 0.796 & 1.602           & 0.865 & 0.129 & 0.795 & 1.403 \\
\bottomrule
\end{tabular}
}
\caption{Occupation classificationm }
\label{tbl:bios}
\vspace{-3pt}
\end{subtable}
    \caption{Results on both tasks, averaged over 5 seeds. Results that have statistically significant difference comparing to the Finetuned results (by Pitman's permutation test, $p < 0.05$) are marked with *.
    +Scrubbing methods that also have statistically significant difference comparing to the Scrubbing results are marked with **.
    $\dagger$ marks debiasing methods that require demographic information. + marks debiasing methods that require demographic information only for hyper-parameter tuning. BLIND successfully reduces bias without using any demographic information.}
\label{tbl:results}
\vspace{-10pt}
\end{table*}

In the main body of the paper, we report accuracy and a representative subset of fairness metrics. The full set of fairness metrics is reported in \Cref{app:results}.

In \Cref{tbl:moji}, we present the results of sentiment classification with racial bias, and in \Cref{tbl:bios}, results on occupation classification with gender bias. As expected, the vanilla fine-tuning baseline yields the best accuracy, but also the worst bias (highest fairness metrics), on both BERT and DeBERTa and on both tasks.

\subsection{Debiasing with Demographic Annotations} 

We first focus on DFL trained with a demographic detector.

\paragraph{Sentiment classification.} 

The auxiliary model is trained to predict race. DFL leads to a statistically significant reduction of bias compared to the fine-tuned baseline in all metrics, with a minor drop in accuracy ($2-3\%$ absolute).
Compared to other methods that use demographic attributes (INLP, RLACE, FairBatch and JTT), DFL maintains better or similar accuracy.
On BERT, it also reduces bias more. On DeBERTa, INLP and RLACE enjoy a greater bias reduction, but suffer a decrease in accuracy ($-14\%$ in INLP, $-2.7\%$ in RLACE), while FairBatch suffers from both a decrease in accuracy and less bias reduction, and JTT does not suffer from accuracy reduction but reduces less bias than DFL.
We conclude that DFL is an effective method to reduce bias in this setting while maintaining high performance on the downstream task. 

\paragraph{Occupation classification.}

When using demographic attributes (here: gender), DFL leads to a statistically significant reduction of bias according to all metrics on both BERT and DeBERTa, with a minor drop in accuracy ($3\%$). In contrast, INLP and RLACE are much less effective in reducing bias in this setting, with no significant difference from the baseline on BERT and only partial improvements on DeBERTa.\footnote{The original INLP and RLACE papers reported better improvements, but they measured bias on an unbalanced test set, while we follow the recommendations in \citet{orgad-belinkov-2022-choose} to use balanced test sets.}
Scrubbing is quite effective in reducing bias while maintaining accuracy, but it achieves a lesser degree of bias reduction than DFL.
When we combine DFL with scrubbing, we find that it achieves an even greater bias reduction, surpassing all other methods, with only a minor accuracy reduction compared to DFL.
JTT reduces bias in only one metric on both models.

Our conclusion is that DFL with demographics is an effective tool for reducing bias, surpassing other methods we compare to.

\subsection{Debiasing without Demographic Annotations.} 

Next, we examine our method when there is no access to demographic attributes (BLIND), using a success detector as a proxy for biased features.

\paragraph{Sentiment classification.}  
Remarkably, we observe a statistically significant reduction of bias compared to the fine-tuned model in BERT and DeBERTa. Reduction in accuracy is minimal, as before. Comparing DFL with and without demographics, the model trained with demographics produces lower fairness metrics, in both BERT and DeBERTa. 
JTT, which also does not use demographics at training, is more effective than BLIND; however, it requires demographics for model selection. Additionally, the control model does not differ statistically significantly from the vanilla model, in both accuracy and fairness metrics.

\paragraph{Occupation classification.} As in the sentiment classification task, debiasing without demographic attributes (BLIND) tends to be less effective than our variant for debiasing using demographic attributes.
Nevertheless, it is still successful in mitigating bias on some of the fairness metrics, even surpassing other methods that use demographic attributes ($-0.69\%$ in sufficiency, compared to $-0.55\%$ for INLP, $-0.23\%$ for RLACE and no significant reduction for JTT), while maintaining a small reduction in performance ($-3\%$).
Once again, the control model results are statistically indistinguishable from those of the baseline.
Additionally, we find that combining BLIND with scrubbing seems to not improve fairness on top of the scrubbing method. Combining BLIND with a method that has access to the bias we wish to remove seems not helpful, at least in this case.

To summarize this part, while the results of DFL without demographic attributes are behind those of DFL that uses attributes and sometimes behind other methods that use attributes, it is encouraging to see a significant reduction in bias in this setting, that is sometimes even more effective than other methods that use demographic attributes. 

\section{Analyses}

\subsection{Performance of the Success Detector}

The success detector achieves an average accuracy of $85\%$ on occupation classification and $76\%$ on sentiment classification with BERT \footnote{see \Cref{app:sd_perf} for results on DeBERTa}. Moreover, we compute its Expected Calibration Error (ECE) and find that it is 0.03 on average for both occupation classification and sentiment classification. These results suggest that the success detector is well-calibrated for both classification tasks and achieves non-trivial accuracy, explaining its effectiveness as a detector for biased samples.

\subsection{Effect of debiasing on internal model representations}

\begin{figure*}[th]
\begin{subfigure}[h]{\columnwidth}
  \includegraphics[width=\columnwidth]{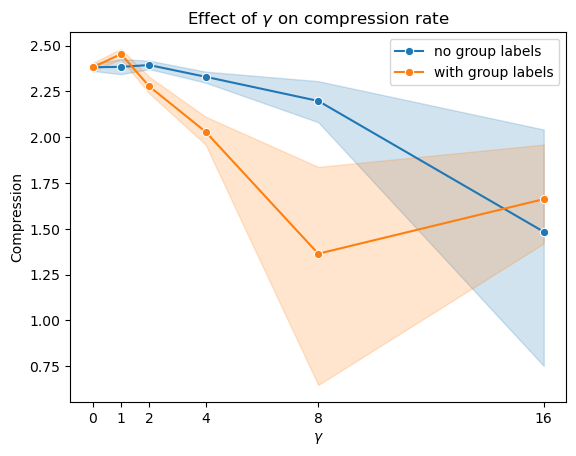}
  \caption{Sentiment classification and race.}
\end{subfigure}
\begin{subfigure}[h]{\columnwidth}
  \includegraphics[width=\columnwidth]{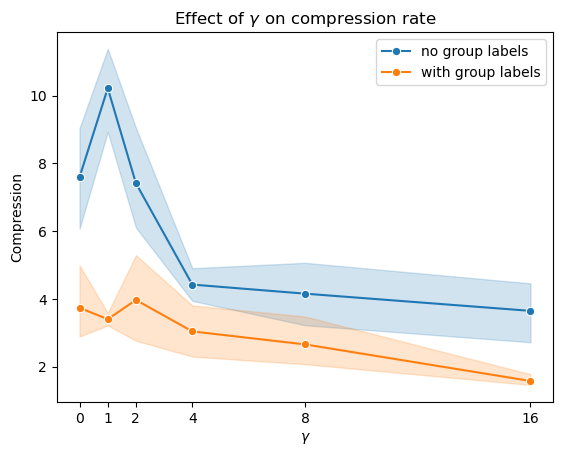}
  \caption{Occupation classification and gender.}
\end{subfigure}
\caption{Effect of $\gamma$ on compression rate of demographic information in internal representations, as extracted from trained models.}
\label{fig:probing}
\end{figure*}

To further understand why BLIND works, we investigate the internal representations of the debiased models. 
Recently, the extractability of gender information from a model's internal representations was found to be correlated with gender bias metrics \cite{orgad-etal-2022-gender}. We therefore pose the following question: How does debiasing with DFL affect the neural representations of demographic information? Here, we focus on BERT.

To answer the above question, we train a probe model, $f_p$, which predicts the protected attribute, either gender or race, from the main model's internal representations: $f_p: g(\mathcal{X}) \rightarrow R^{|Z|}$. We then report the ease at which the probe performs this task using compression, measured by a minimum description length (MDL) probing  \cite{voita-titov-2020-information}.\footnote{MDL probes address the problem with looking at a probe's accuracy as measure of information, which has been found to be misleading due to memorization and other factors \cite{hewitt-liang-2019-designing, probing-belinkov}.} Internal representations with a higher compression indicate more accessible gender or racial information. A detailed description on the implementation can be found in \Cref{app:mdl}.

Figure \ref{fig:probing} presents the probing results for both tasks and on the two variations of debiasing: with and without demographic attributes. Even though some $\gamma$s are noisy, there is a clear trend that the accessibility of demographic information decreases as $\gamma$ increases. Surprisingly, applying BLIND caused the models to encode less demographics information even without information about the protected attributes. This may explain why BLIND is successful in reducing bias metrics associated with these demographics, as well as suggesting that other hidden characteristics may also be affected by this debiasing process. These results align with \citet{orgad-etal-2022-gender}, who found that the extractability of gender in internal representations correlate positively with various fairness metrics. However, our results are different from those of \citet{mendelson-belinkov-2021-debiasing}, who found a negative correlation between robustness and biased features extractability. \citeauthor{mendelson-belinkov-2021-debiasing} explicitly modeled biased features for their debiasing process, whereas we use demographics or success as proxies for biased features, which might explain the difference.

\subsection{Bias detectors comparison: with and without demographics}

Recall that our method penalizes samples for which the detector assigned a high probability to the correct label. The results indicated that using a demographic detector (race/gender) was more effective than using a success detector. The two detector models penalize samples differently, so we wish to understand how the two differ and where they agree. We compute the probability that each model provided to the correct label (details on computation in \Cref{app:comparison_computation}).\footnote{The correct labels for the success detector are the main model's success or failure for each sample. The correct labels for the race/gender detector are the protected attributes $z_i$.} Here we present our analysis of BERT. Results for DeBERTa are similar (\Cref{app:relationship}).

\begin{table}[t]
\centering
\adjustbox{max width=\columnwidth}{%
\begin{tabular}{l|lll}
\toprule
& \multirow{2}{*}{\textbf{\shortstack[l]{Success \\ detector}}} & \multirow{2}{*}{\textbf{\shortstack[l]{Demographics \\ detector only}}} & \multirow{2}{*}{\textbf{\shortstack[l]{Both}}} \\
\\
\midrule
\textbf{Sentiment} & 94\%/22\%        & 4\%/55\%          & 69\%/15\% \\
\midrule
\textbf{Occupation}   & 95\%/33\%        & 5\%/66\%          & 87\%/32\% \\
\bottomrule
\end{tabular}
}
\caption{
Percentage of success/failure samples penalized by our detectors, out of all  success/failure samples. Left column: samples penalized by the success detector, middle column: by the demographics detector and \textbf{not} by the success detector, right column: by both models.}
\vspace{-15pt}
\label{tbl:cartography_bert}
\end{table}

\Cref{tbl:cartography_bert} summarizes which samples each detector penalizes, defined as samples for which the detector assigns a probability above $0.5$. We divide the samples into two classes, depending on whether the main model classified them correctly or not.\footnote{We also analyzed an alternative sample division: AAE/not AAE and female/male, but we found that both detectors down-weighted these groups equally.} The table shows the percentage of samples that are being penalized, out of all samples in this class.

We first note that the success detector (left column) is mostly penalizing samples where the main model is correct (94\% and 95\% in sentiment and occupation tasks, respectively), and much less samples where the main model is wrong (22\% and 33\%). Thus, the success detector reduces weights on samples the model has already learned and classified correctly, which could correspond to easier samples that contain more bias. This reduces the overall bias by preventing the main model from over-fitting to these samples.

Looking at samples that both detectors penalize (right column), we observe that they are mostly samples which the main model succeeds on (69\% for sentiment and 87\% for occupation), suggesting their importance for debiasing. However, when observing what samples the demographics detector penalized but the success detector did not (middle column), we find many failure samples (55\% for sentiment and 66\% for occupation). In our experiments, our method with demographics mitigated bias better than the one without. The gap between the methods might be because failure samples are less penalized by the success detector, since the success detector fails to correctly classify these samples. Better debiasing might be achieved by detecting failures in the main model more effectively, perhaps by using a stronger success detector.

\section{Prior Work}

Studies suggests a variety of ways for debiasing NLP models from social biases on downstream tasks, such as preprocessing the training data \cite{dearteaga-bios, zhao-etal-2018-gender, han2021balancing}, modifying the training process \cite{elazar-goldberg-2018-adversarial, shen-etal-2022-optimising}, or applying post-hoc methods to neural representations of a trained model \cite{ravfogel-etal-2020-null, pmlr-v162-ravfogel22a, iskander2023shielded}. All these methods, however, require that we define the bias we wish to operate upon, for example, gender bias. Additionally, many of these methods require demographic annotations per data instance, such as the gender of the writer or the subject of the text. \citet{webster2020measuring} is an exception, since it performs gender-debiasing by modifying dropout parameters. Another exception is JTT \cite{pmlr-v139-liu21f}, which improved worst-group errors by training a model twice: first a standard ERM model, then a second model that upweights training samples misclassified by the first model. The authors of these studies chose hyper-parameters based on fairness metrics they wanted to optimize, while we choose our hyper-parameters without explicitly measuring fairness metrics. To our knowledge, this is the first study to mitigate social biases in NLP without assuming any prior knowledge.

Other studies have focused on improving NLP models robustness without prior knowledge of bias issues, but without considering social bias. \citet{utama-etal-2020-towards} and \cite{sanh2020learning} tackled dataset biases (a.k.a annotation artifacts) in natural language understanding tasks, by training a weak learner to identify biased samples and down-weighting their importance. Weak learners are either trained on a random subset of the dataset or have a smaller capacity.

Regarding social fairness without demographics, \citet{lahoti-fairness-without-demog} proposed adversarially reweighted learning, where an adversarial model is trained to increase the total loss by re-weighting the training samples. They used tabular, non-textual data in their experiments. We consider non-adversarial methods since adversarial training is known to be difficult to stabilize. \citet{hashimoto2018fairness} proposed a method for minimizing the worst case risk over all distributions close to the empirical distribution, without knowledge of the identity of the group. \citet{coston2019} considered fairness in unsupervised domain adaptation, where the source or target domain do not have demographic attributes, by proposing weighting methods. \citet{han-etal-2021-decoupling} proposed debiasing via adversarial training with only a small volume of protected labels.

Focal loss \cite{lin2017focal} was proposed as a method to address class imbalances by down-weighting loss associated with well-classified samples. \citet{rajivc2022using} proposed using the original focal loss \cite{lin2017focal} to improve robustness in natural language inference, leading to improved out-of-distribution accuracy. Debiased Focal Loss (DFL) \cite{karimi-mahabadi-etal-2020-end} is a variant of focal loss proposed to improve natural language understanding models on out-of-distribution data.

\section{Discussion and Conclusion}

Even though BLIND led to bias reduction, it was less effective than our method that used demographic annotations. Analysis showed that the success detector is less accurate at classifying samples that fail the main model. Additionally, the success detector might be less focused than demographic-based methods, but it might mitigate biases we have not identified and cannot measure without annotations. Thus, it would be interesting to see how BLIND works on intersectional biases.

In sentiment analysis, BLIND reduced bias less than JTT, which also does not use demographics for training, but does for hyper-parameter search. However, JTT was ineffective on the occupation classification task, while BLIND was effective for both tasks. The two tasks differ significantly, as well as their data. For BERT and DeBERTa, pretraining data is closer to biographies than tweets, so perhaps training for longer is beneficial for the tweets data used in the sentiment classification task, and repeating samples in the training set (as in JTT) is similar to training for more steps. In any case, BLIND proved more reliable and generalizable in reducing bias.

Our method has the potential for broader applications beyond demographic biases. While our primary focus was on mitigating demographic biases, the approach can be adapted to address other types of biases by identifying relevant proxy indicators via the success detector.

To summarize, we demonstrated the reduction of racial and gender biases of NLP classification models, without any prior knowledge of those biases in the data. These results suggest that we can potentially apply BLIND to any dataset, which makes bias reduction a much more feasible endeavor and applicable to real-world scenarios.

\section{Limitations}

One limitation of this study is its scope, which covers two downstream tasks and two types of demographics (race and gender). The binary gender definition we used excludes other genders that do not fall under male or female. In the case of race, we explored only African American race (proxied by African American English), which excludes biases related to other races, and is a US-centric view of racial bias. We did not investigate other types of bias, such as religious bias. Furthermore, our method was tested on datasets with short texts, and it is unclear how it will perform on longer texts. The experiments were conducted on datasets in English, and it is unclear how our method will work on languages that are morphologically rich.

\section{Ethics Statement}

Through this study, we aim to reduce the barriers of data collection in the effort of mitigating bias. In situations where demographic information is not available at all, or where its use could cause privacy concerns, this method may be especially useful. As with other bias mitigation methods, applying BLIND to the training process might create a false sense of confidence in the model's bias, but as we target scenarios without demographics, the risk is greater as it may be harder to discover cases where bias remains. We encourage practitioners of NLP who use BLIND to identify potential biases and harms experienced by individuals using their system, and to define their fairness metrics accordingly. In order to verify if the system is working as expected according to the predefined fairness metrics, we encourage collecting a small validation set with demographic annotations.

\section*{Acknowledgements}
This research was supported by the ISRAEL SCIENCE FOUNDATION (grant No.\ 448/20) and by an Azrieli Foundation Early Career Faculty Fellowship.

\bibliography{anthology,custom}
\bibliographystyle{acl_natbib}

\appendix

\section{Connection of DFL to adversarial training}
\label{app:adversarial_learning}

As adversarial learning is a natural competing baseline, we attempted to apply adversarial training to our task setting at an initial step of the research, but found it to be highly unstable and impractical, which corroborates findings in the literature \cite{ganin2016domain, grand-belinkov-2019-adversarial}. We also examined the work of \citet{elazar-goldberg-2018-adversarial}, who utilized adversarial learning for removing demographic attributes from text inputs. However, they caution against relying on adversarial removal for achieving fairness, as their results indicated that demographic information could still be extracted by classifiers of the same architecture.
To address the issue of attribute leakage more effectively, subsequent work \cite{ravfogel-etal-2020-null, pmlr-v162-ravfogel22a} proposed other methods that address demographic attribute leakage better than adversarial training, which we compare our method with.

\section{DTO}
\label{app:dto}

DTO \citet{han2021balancing} is measured as the L2-distance from a utopia point, $(\mathtt{max_accuracy}, 0)$, where $\mathtt{max_accuracy}$ is the maximum accuracy achieved on the task in our experiments. $(\mathtt{accuracy}, \mathtt{fairness})$ are the candidate points. We computed $\mathtt{fairness}$ by averaging the various fairness metrics measured in this study. In a practical application, the definition of the candidate points and the utopia points should reflect the application's current needs and priorities.

\section{Implementation Details}
\label{app:training}

\subsection{Fairness Metrics}
\label{app:metrics}

For measuring the statistical fairness metrics (independence, separation, sufficiency), we used the \texttt{fairness} library, implemented by AllenNLP (\url{https://github.com/allenai/allennlp}).

\paragraph{FairBatch fairness metrics.} FairBatch considers three fairness metrics: equal opportunity, equalized odds, and demographic parity. Equal opportunity is closely related to the metric Independence, Equalized odds is closely related to Separation and equalized opportunity is closely related to TPR gap. In the main results table (Table \ref{tbl:moji}) we present the variation that achieved the best metrics we display. However, we found very little difference between the results of the different variations of FairBatch.

\subsection{Training and Evaluation}

For the DFL loss, increasing the temperature $t$ increases the smoothness of the softmax result, resulting in less radical regularization. In each experiment, we grid-search $\gamma \in \{1, 2, 4, 8, 16\}$ and $t \in \{1, 2, 4, 8\}$ using the validation set. 

We trained BERT \cite{devlin2018bert} and DeBERTa-v1 \cite{he2020deberta}  models, with one linear classification layer on top of them. For BERT we used the $\texttt{bert-base-uncased}$ model by Huggingface, which has 110M parameters and for DeBERTa we used the $\texttt{microsoft/debeta-base}$ from the Huggingface library, which has 1.5B parameters. We trained each model for 10 epochs---which took around 5-6 hours for the occupation classication task and 3-4 for the sentiment classification task---and saved the best epoch based on the validation accuracy.

We used the following GPUs for training BERT: Geforce RTX 2080 Ti, TITAN Xp, GeForce GTX 1080 Ti, and the following GPUs for training DeBERTa: A40, RTX A4000. Each experiment was run using 5 different seeds: 0, 5, 26, 42, 63. These seeds were used to anchor the model's initialization, the data split, and any other randomness in the code, and are considered as an input to our released scripts. We used a 65-10-25 training-validation-test split ratio for all tasks. Training was done with a learning rate of $5e-5$ and an AdamW \cite{loshchilov2018decoupled} optimizer. Both models were trained by passing the biography / tweet through the transformer model, obtaining the top [CLS] token representation and feeding it into the classification layer. The entire model was fine-tuned end-to-end to optimize the cross entropy loss, while in the DFL setting we added a weighting component. The demographics/success detector's architecture is a single linear classification layer optimized to solve the appropriate classification task. It was trained with a learning rate of $1e-3$ and an Adam \cite{kingma2015adam} optimizer.

\paragraph{Sentiment Classification.} Our setup followed the settings of \citet{ravfogel-etal-2020-null}, where we controlled how biased the training data is. We used a subset of the original dataset with 100K samples, and our training data was imbalanced such that the ``happy'' sentiment class was composed of 70\% AAE and 30\% MUSE, while the ``sad'' sentiment was composed of 70\% MUSE and 70\% AAE. The dataset was overall balanced: 50\% ``sad''/``happy'' and 50\% AAE/MUSE.

\paragraph{Occupation Classification.} We trained our models on the entire dataset, without any distribution modifications.

For both tasks, data for validation and testing was balanced such that each class had the same demographic distribution.

\subsection{Baselines and Competitive Systems}

We ran INLP and RLACE on model representations extracted from a finetuned model without DFL training. For INLP and RLACE, we used the implementation of the authors and the same hyper-parameters. For JTT, since the datasets were different, we provide our own implementation but searched over the same hyper-parameters, which are $\lambda_{up}\in \{4, 5, 6\}$ (the rate of multiplication of repeated instances) and $T=\{1, 2\}$ (the epoch of the model used to calculate failed instances). We ran the scrubbing algorithm provided by the code of \citeauthor{dearteaga-bios}. FairBatch was ran using the code provided in the paper, for a training period of 10 epochs where the best checkpoint is chosen as the epoch with the best fairness metric being optimized.

\subsection{MDL probing: implementation details}
\label{app:mdl}

Our MDL probe \cite{voita-titov-2020-information} is based on the implementation by \citet{mendelson-belinkov-2021-debiasing}. The linear probe is trained with a batch size of 16 and a learning rate of 1e-3 in all experiments. We used the following data accumulation fractions to train the probe: 2.0\%, 3.0\%, 4.4\%, 6.5\%, 9.5\%, 14.0\%, 21.0\%, 31.0\%, 45.7\%, 67.6\%, 100\%.

\subsection{Bias detectors comparison: implementation details}
\label{app:comparison_computation}

We wished to calculate for each model, the probability it assigned to the right label. To do that, we load the checkpoint of a BLIND-trained model we are interested in, and then compute the main model's predictions and the hidden representations of each sample from a BLIND-trained model. Following that, we calculate the success detector's and the demographics detector's predictions on the extracted representations. For the demographics detector, we train another model for predicting demographics from the extracted representations, to simulate how a demographics detector would behave on the same representations.

\section{Full results}
\label{app:results}

The full results can be found in Tables \ref{tbl:full_results_moji_bert}, \ref{tbl:full_results_moji_deberta}, \ref{tbl:full_results_bios_bert} and \ref{tbl:full_results_bios_deberta}.

\section{Effect of $\gamma$}

\label{app:gamma}
\begin{figure*}[th]
\begin{subfigure}[h]{.5\columnwidth}
  \includegraphics[width=\columnwidth]{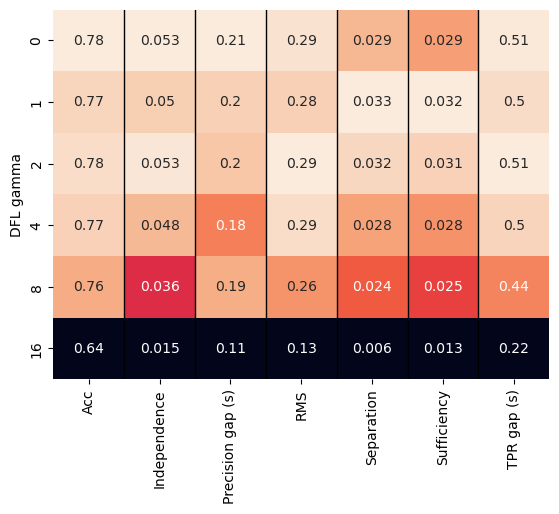}
  \caption{$t=1$}
\end{subfigure}
\begin{subfigure}[h]{.5\columnwidth}
  \includegraphics[width=\columnwidth]{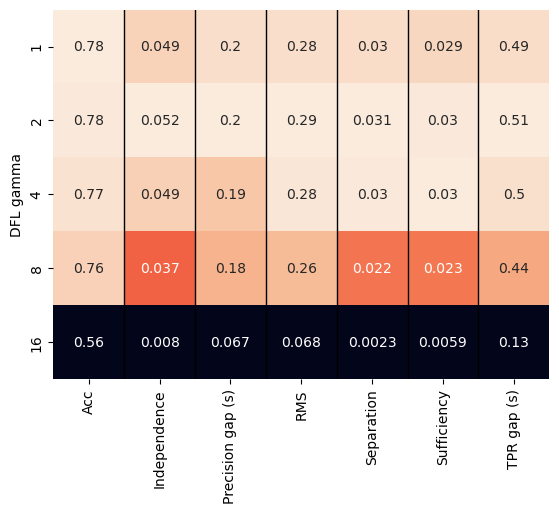}
  \caption{$t=2$}
\end{subfigure}
\begin{subfigure}[h]{.5\columnwidth}
  \includegraphics[width=\columnwidth]{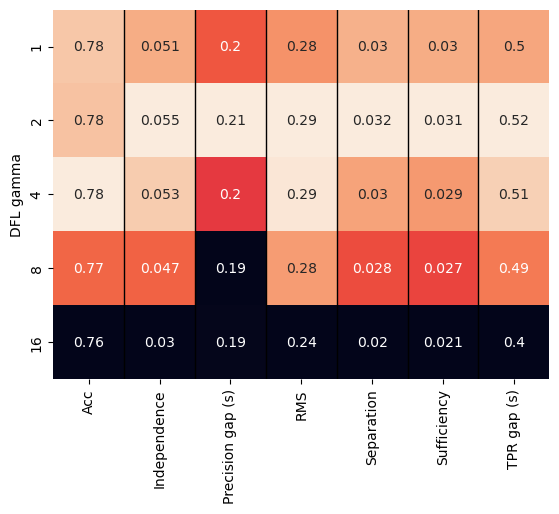}
  \caption{$t=4$}
\end{subfigure}
\begin{subfigure}[h]{.5\columnwidth}
  \includegraphics[width=\columnwidth]{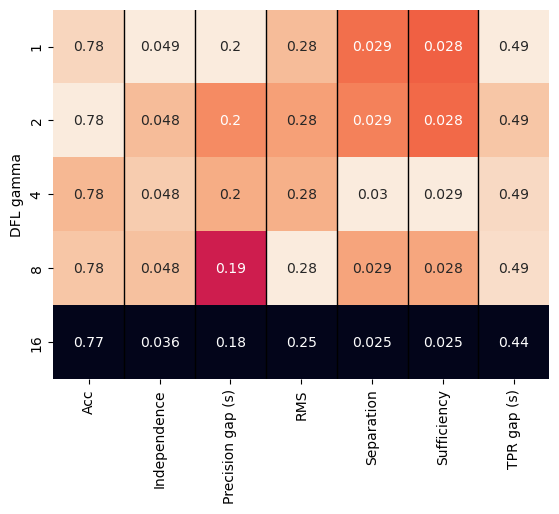}
  \caption{$t=8$}
\end{subfigure}
\begin{subfigure}[h]{.19\textwidth}
  \includegraphics[width=\columnwidth]{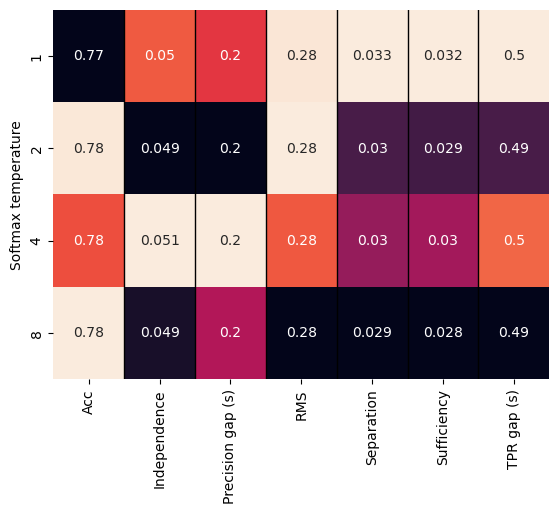}
  \caption{$\gamma=1$}
\end{subfigure}
\begin{subfigure}[h]{.19\textwidth}
  \includegraphics[width=\columnwidth]{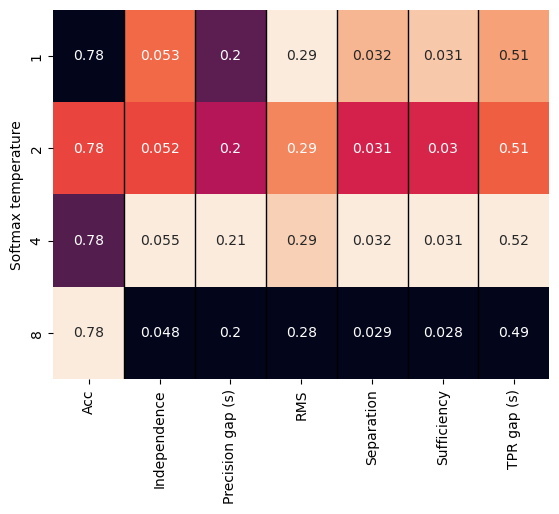}
  \caption{$\gamma=2$}
\end{subfigure}
\begin{subfigure}[h]{.19\textwidth}
  \includegraphics[width=\columnwidth]{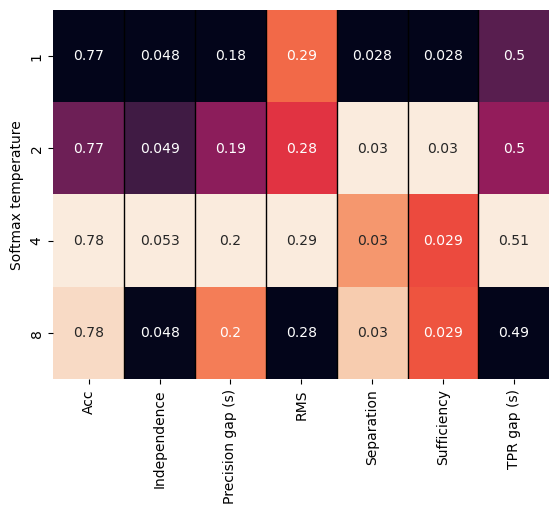}
  \caption{$\gamma=4$}
\end{subfigure}
\begin{subfigure}[h]{.19\textwidth}
  \includegraphics[width=\columnwidth]{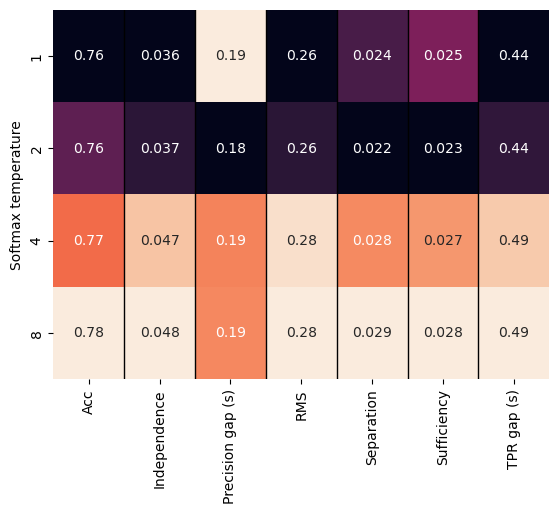}
  \caption{$\gamma=8$}
\end{subfigure}
\begin{subfigure}[h]{.19\textwidth}
  \includegraphics[width=\columnwidth]{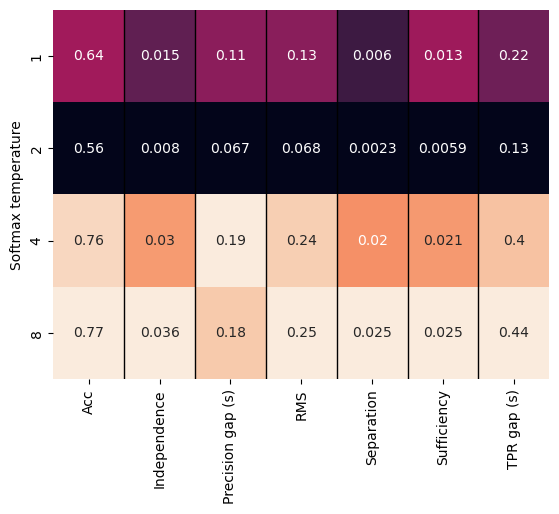}
  \caption{$\gamma=16$}
\end{subfigure}
\caption{Sentiment classification, BERT. effect of $\gamma$ across different softmax temperatures ($t$), and effect of different softmax temperature across different $\gamma$.}
\label{fig:moji_bert_gamma}
\end{figure*}

\begin{figure*}[th]
\begin{subfigure}[h]{.5\columnwidth}
  \includegraphics[width=\columnwidth]{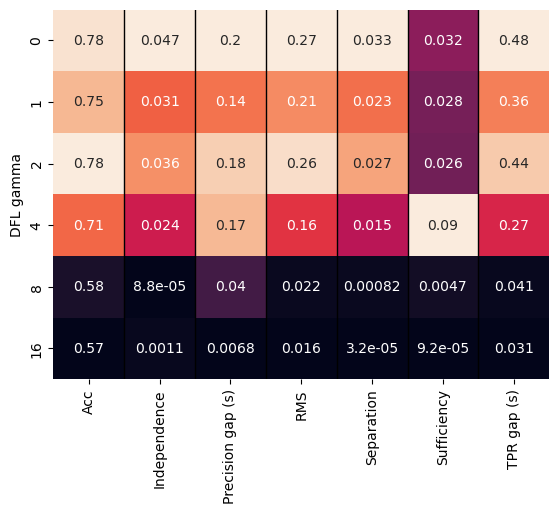}
  \caption{$t=1$}
\end{subfigure}
\begin{subfigure}[h]{.5\columnwidth}
  \includegraphics[width=\columnwidth]{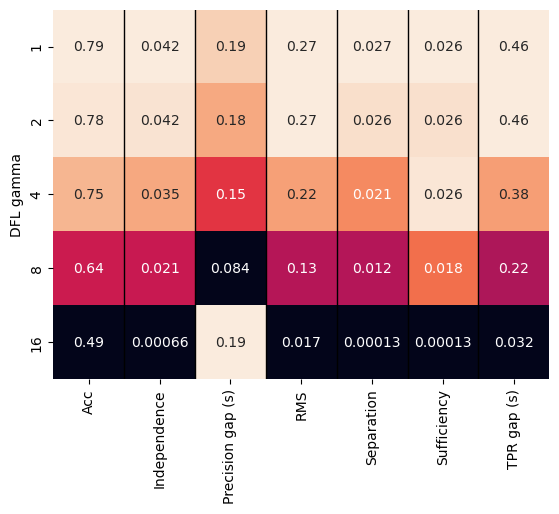}
  \caption{$t=2$}
\end{subfigure}
\begin{subfigure}[h]{.5\columnwidth}
  \includegraphics[width=\columnwidth]{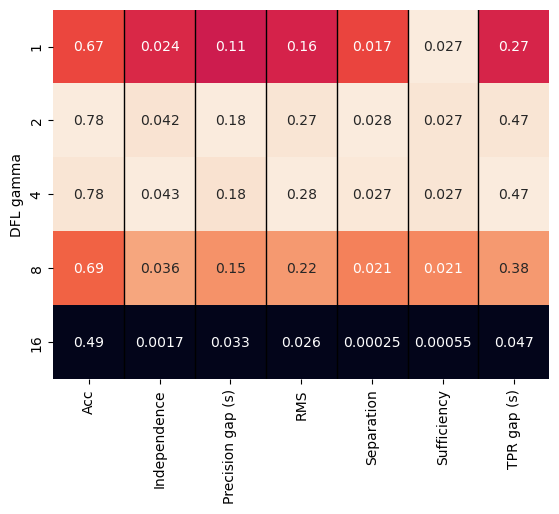}
  \caption{$t=4$}
\end{subfigure}
\begin{subfigure}[h]{.5\columnwidth}
  \includegraphics[width=\columnwidth]{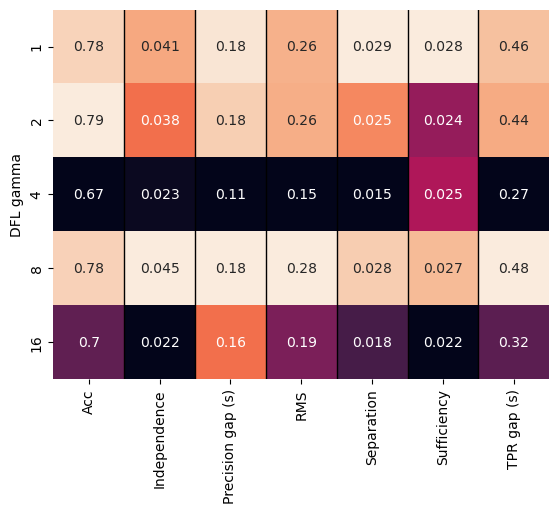}
  \caption{$t=8$}
\end{subfigure}
\begin{subfigure}[h]{.19\textwidth}
  \includegraphics[width=\columnwidth]{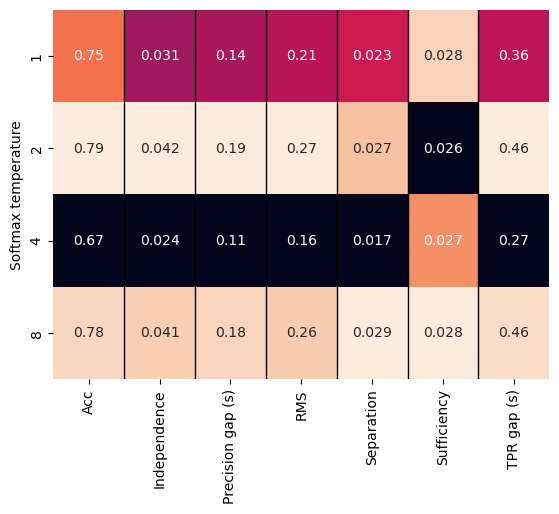}
  \caption{$\gamma=1$}
\end{subfigure}
\begin{subfigure}[h]{.19\textwidth}
  \includegraphics[width=\columnwidth]{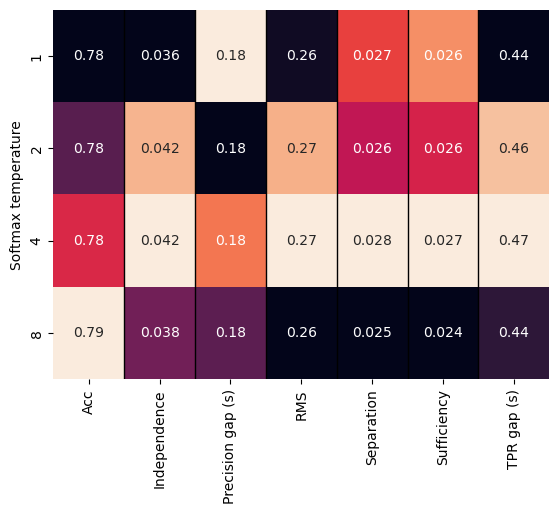}
  \caption{$\gamma=2$}
\end{subfigure}
\begin{subfigure}[h]{.19\textwidth}
  \includegraphics[width=\columnwidth]{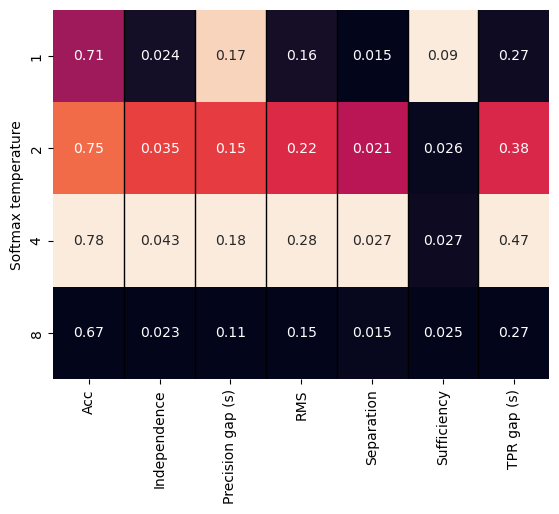}
  \caption{$\gamma=4$}
\end{subfigure}
\begin{subfigure}[h]{.19\textwidth}
  \includegraphics[width=\columnwidth]{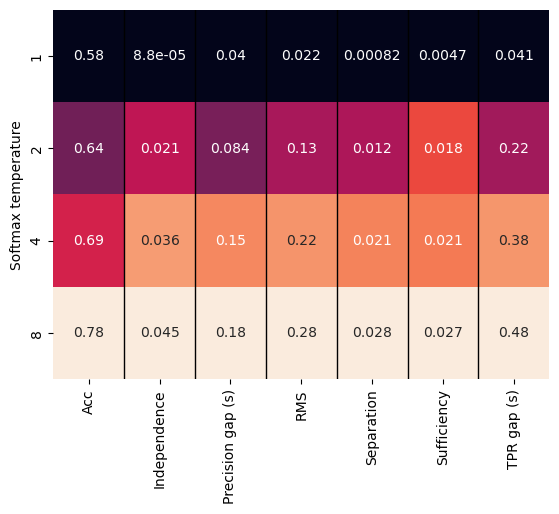}
  \caption{$\gamma=8$}
\end{subfigure}
\begin{subfigure}[h]{.19\textwidth}
  \includegraphics[width=\columnwidth]{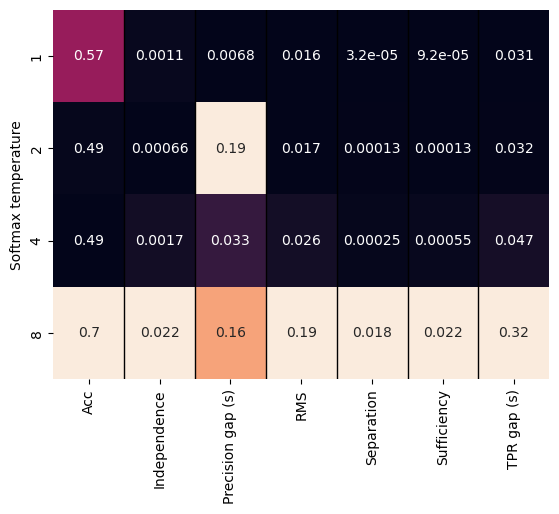}
  \caption{$\gamma=16$}
\end{subfigure}
\caption{Sentiment classification, DeBERTa. effect of $\gamma$ across different softmax temperatures ($t$), and effect of different softmax temperature across different $\gamma$.}
\label{fig:moji_deberta_gamma}
\end{figure*}

\begin{figure*}[th]
\begin{subfigure}[h]{.5\columnwidth}
  \includegraphics[width=\columnwidth]{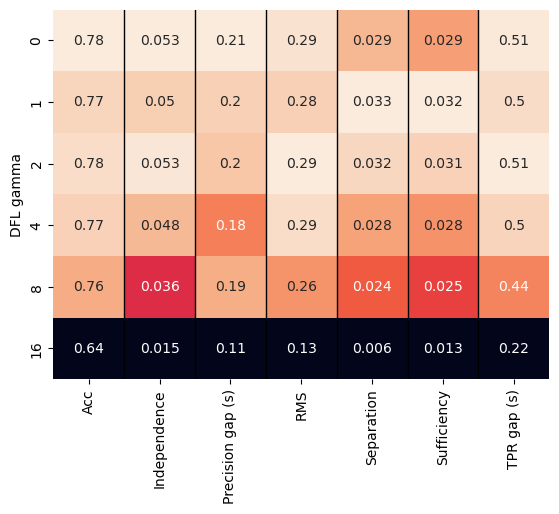}
  \caption{$t=1$}
\end{subfigure}
\begin{subfigure}[h]{.5\columnwidth}
  \includegraphics[width=\columnwidth]{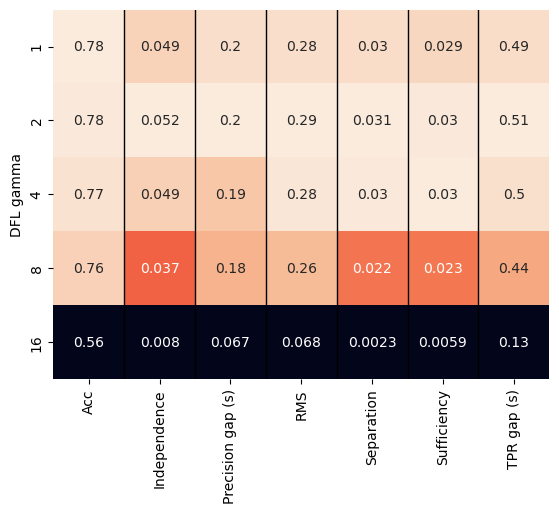}
  \caption{$t=2$}
\end{subfigure}
\begin{subfigure}[h]{.5\columnwidth}
  \includegraphics[width=\columnwidth]{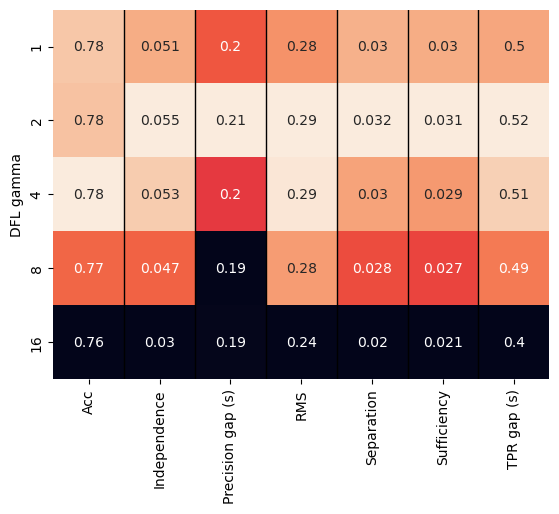}
  \caption{$t=4$}
\end{subfigure}
\begin{subfigure}[h]{.5\columnwidth}
  \includegraphics[width=\columnwidth]{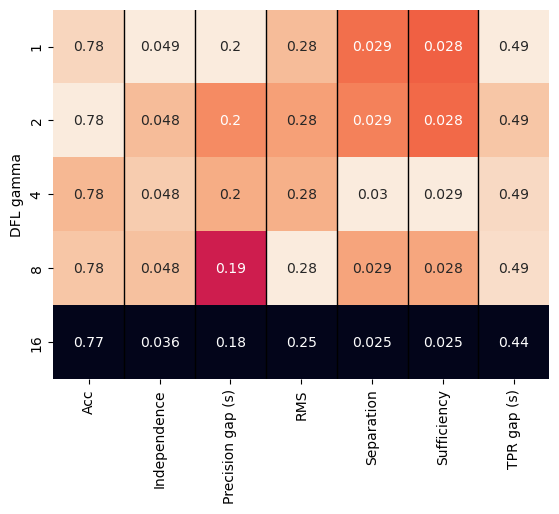}
  \caption{$t=8$}
\end{subfigure}
\begin{subfigure}[h]{.19\textwidth}
  \includegraphics[width=\columnwidth]{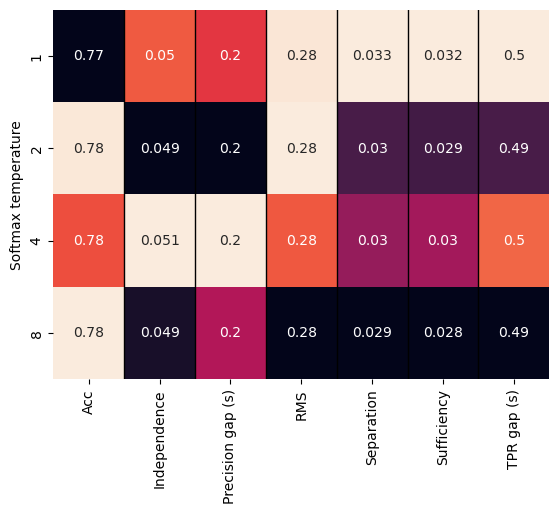}
  \caption{$\gamma=1$}
\end{subfigure}
\begin{subfigure}[h]{.19\textwidth}
  \includegraphics[width=\columnwidth]{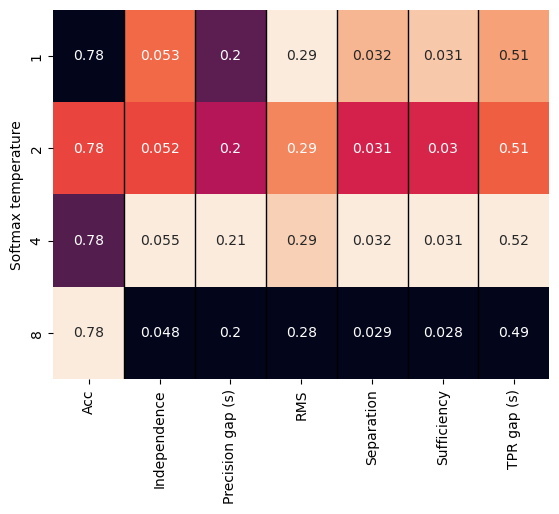}
  \caption{$\gamma=2$}
\end{subfigure}
\begin{subfigure}[h]{.19\textwidth}
  \includegraphics[width=\columnwidth]{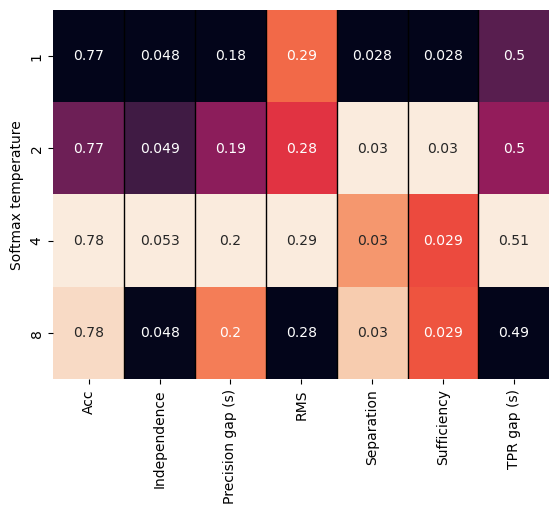}
  \caption{$\gamma=4$}
\end{subfigure}
\begin{subfigure}[h]{.19\textwidth}
  \includegraphics[width=\columnwidth]{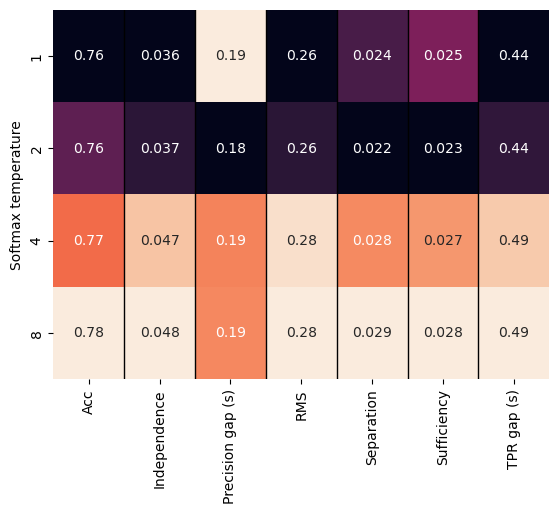}
  \caption{$\gamma=8$}
\end{subfigure}
\begin{subfigure}[h]{.19\textwidth}
  \includegraphics[width=\columnwidth]{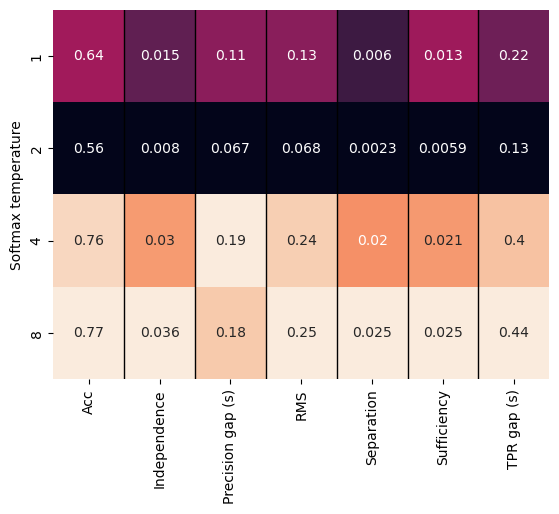}
  \caption{$\gamma=16$}
\end{subfigure}
\caption{Occupation classification, BERT. effect of $\gamma$ across different softmax temperatures ($t$), and effect of different softmax temperature across different $\gamma$.}
\label{fig:bios_bert_gamma}
\end{figure*}

\begin{figure*}[th]
\begin{subfigure}[h]{.5\columnwidth}
  \includegraphics[width=\columnwidth]{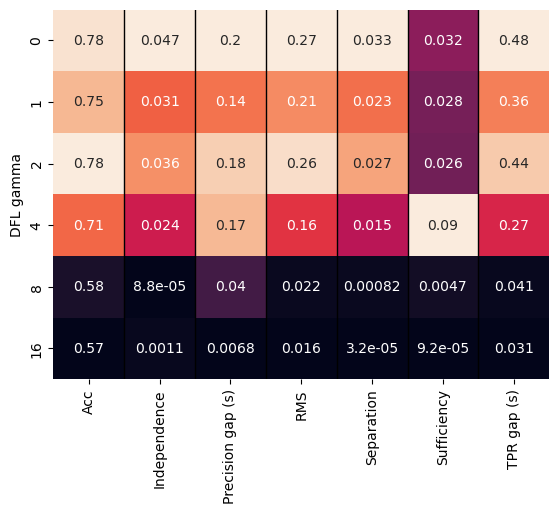}
  \caption{$t=1$}
\end{subfigure}
\begin{subfigure}[h]{.5\columnwidth}
  \includegraphics[width=\columnwidth]{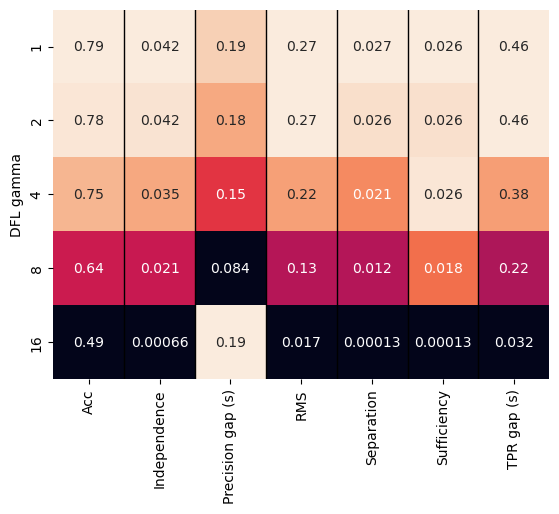}
  \caption{$t=2$}
\end{subfigure}
\begin{subfigure}[h]{.5\columnwidth}
  \includegraphics[width=\columnwidth]{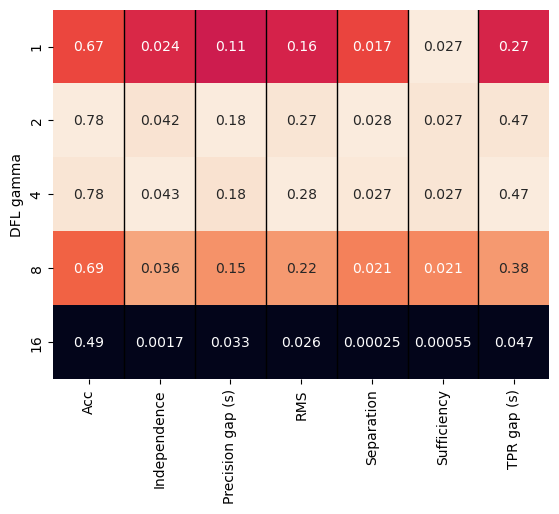}
  \caption{$t=4$}
\end{subfigure}
\begin{subfigure}[h]{.5\columnwidth}
  \includegraphics[width=\columnwidth]{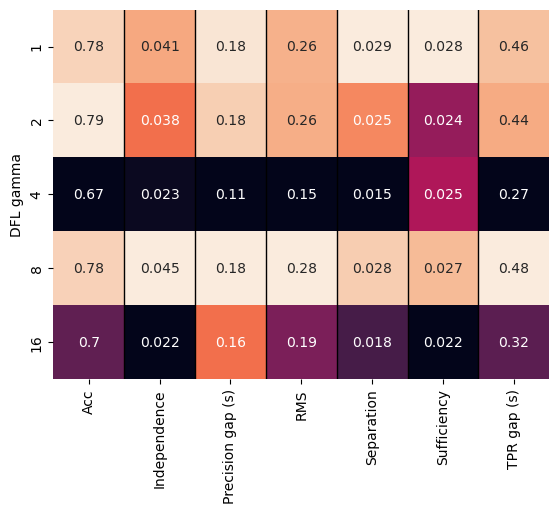}
  \caption{$t=8$}
\end{subfigure}
\begin{subfigure}[h]{.19\textwidth}
  \includegraphics[width=\columnwidth]{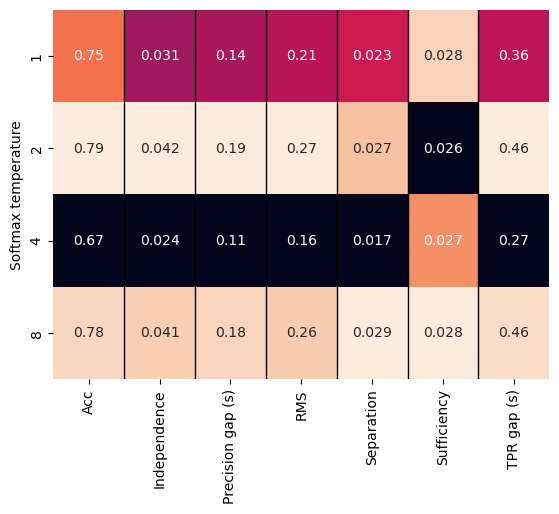}
  \caption{$\gamma=1$}
\end{subfigure}
\begin{subfigure}[h]{.19\textwidth}
  \includegraphics[width=\columnwidth]{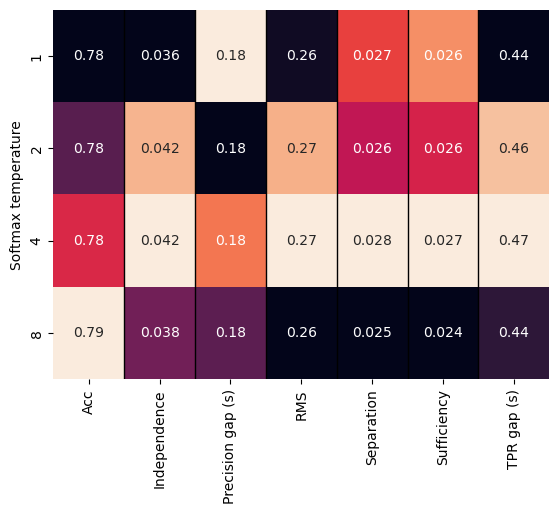}
  \caption{$\gamma=2$}
\end{subfigure}
\begin{subfigure}[h]{.19\textwidth}
  \includegraphics[width=\columnwidth]{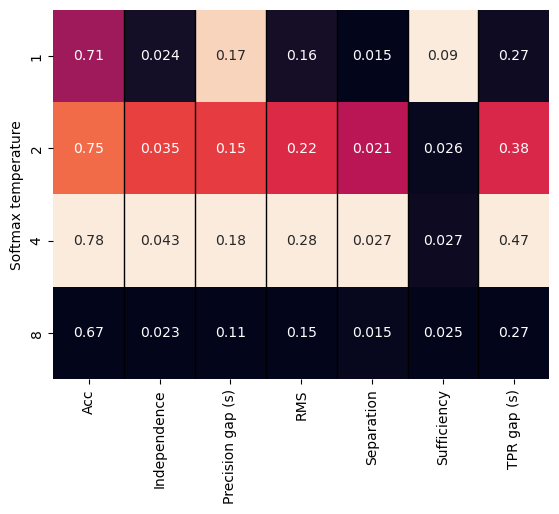}
  \caption{$\gamma=4$}
\end{subfigure}
\begin{subfigure}[h]{.19\textwidth}
  \includegraphics[width=\columnwidth]{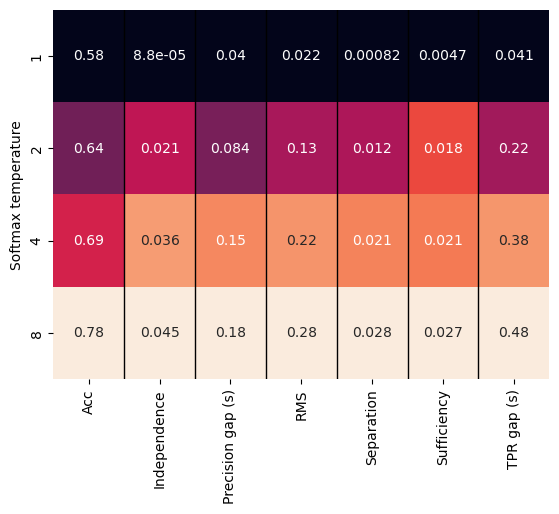}
  \caption{$\gamma=8$}
\end{subfigure}
\begin{subfigure}[h]{.19\textwidth}
  \includegraphics[width=\columnwidth]{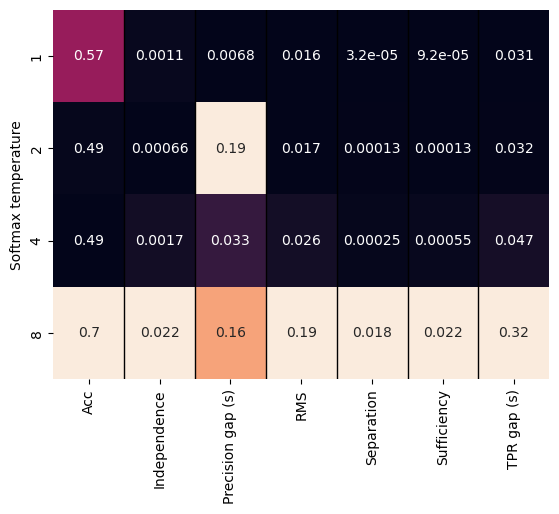}
  \caption{$\gamma=16$}
\end{subfigure}
\caption{Occupation classification, DeBERTa. effect of $\gamma$ across different softmax temperatures ($t$), and effect of different softmax temperature across different $\gamma$.}
\label{fig:bios_deberta_gamma}
\end{figure*}

Recall that the DFL loss (\Cref{eqn:dfl_formulation}) uses $\gamma$ to control how much importance to give to a biased sample; the higher $\gamma$, the less weight a biased sample receives in the loss, which should result in a more biased model. 
Indeed, Figures \ref{fig:moji_bert_gamma}, \ref{fig:moji_deberta_gamma}, \ref{fig:bios_bert_gamma} and \ref{fig:bios_deberta_gamma} show that as $\gamma$ increases (moving from top to bottom on the heatmaps), the bias metrics and accuracy both tend to decrease.

\paragraph{Choosing the hyper-parameters blindly.} As the figures show, increasing $\gamma$ causes the debiasing effect to be more aggresive, until a point that it is collapsing and unable to train (very low accuracy). Increasing the temperature helps balancing this process, where on the higher gammas a lower temperature mean more aggressive debiasing and thus less bias. Based on this analysis, we conclude that it is possible to make the model selection based on $\gamma$ and $t$ alone, using the following logic: choose the highest $\gamma$ and the lowest $t$ for which the accuracy is above a tolerance threshold.

\begin{table}[h]
\centering
\adjustbox{max width=\columnwidth}{%
\begin{tabular}{llrrrr}
\toprule
\textbf{Model} & \textbf{Task} & \multicolumn{2}{l}{\textbf{Accuracy}} & \multicolumn{2}{l}{\textbf{ECE}} \\
\cmidrule{1-6}
             &      & Mean &  Std & Mean &  Std \\
\midrule
BERT & Occupation & 0.85 & 0.00 & 0.03 & 0.02 \\
             & Sentiment & 0.76 & 0.01 & 0.03 & 0.01 \\
DeBERTA & Occupation & 0.86 & 0.01 & 0.03 & 0.01 \\
             & Sentiment & 0.81 & 0.01 & 0.03 & 0.01 \\
\bottomrule
\end{tabular}
}
\caption{Performance and calibration of BLIND's success detector on both tasks.}
\label{tbl:calibration}
\end{table}
\section{Performance of Success Detector}
\label{app:sd_perf}

Table \ref{tbl:calibration} presents the full performance and calibration information of the success detector in both tasks.

\newpage
\section{Bias detectors comparison: with and without demographics}
\label{app:relationship}
\begin{table}[t]
\centering
\adjustbox{max width=\columnwidth}{%
\begin{tabular}{l|lll}
\toprule
& \multirow{2}{*}{\textbf{\shortstack[l]{Success \\ detector}}} & \multirow{2}{*}{\textbf{\shortstack[l]{Demographics \\ detector only}}} & \multirow{2}{*}{\textbf{\shortstack[l]{Both}}} \\
\\
\midrule
\textbf{Sentiment} & 95\%/13\%        & 4\%/71\%          & 82\%/11\% \\
\midrule
\textbf{Occupation}   & 94\%/36\%        & 6\%/59\%          & 84\%/34\% \\
\bottomrule
\end{tabular}
}
\caption{
Percentage of success/failure samples penalized by our detectors, out of all  success/failure samples.}
\label{tbl:cartography_deberta}
\end{table}
Table \ref{tbl:cartography_deberta} presents the results of bias detectors comparison for DeBERTa.

\begin{table*}[ht]
\adjustbox{max width=\textwidth}{%

\begin{tabular}{lllllllll}
\toprule
{} & \multicolumn{8}{c}{BERT} \\
\cmidrule(lr){2-9}
{} & Acc $\uparrow$ & TPR (s) &  $\text{TPR}_{RMS}$ $\downarrow$ & $\text{FPR}_s$ $\downarrow$ & $\text{Prec}_s$ $\downarrow$ & Indep $\downarrow$ & Sep $\downarrow$ & Suff $\downarrow$ \\
\midrule
\textbf{Finetuned} \\ 
Mean & 0.779 & 0.472 & 0.267 & 0.472 & 0.189 & 0.045 & 0.028 & 0.027 \\
Std & 0.006 & 0.034 & 0.019 & 0.034 & 0.016 & 0.007 & 0.003 & 0.003 \\
\textbf{INLP} $\dagger$ &  \\
Mean & 0.756 & 0.334 & 0.196 & 0.334 & \textbf{0.144} & 0.022 & 0.014 & 0.014 \\
Std & 0.023 & 0.077 & 0.045 & 0.077 & 0.035 & 0.009 & 0.008 & 0.008 \\
\textbf{RLACE} $\dagger$ & \\
Mean & 0.735 & 0.247 & 0.142 & 0.247 & 0.230 & 0.010 & 0.003 & 0.008 \\
Std & 0.003 & 0.020 & 0.013 & 0.020 & 0.016 & 0.003 & 0.001 & 0.002 \\
\textbf{FairBatch-DP} $\dagger$ \\
Mean & 0.761 & 0.334 & 0.210 & 0.334 & 0.199 & 0.019 & 0.015 & 0.016 \\
Std & 0.009 & 0.020 & 0.013 & 0.020 & 0.009 & 0.003 & 0.003 & 0.002 \\
\textbf{FairBatch-Eqodds} $\dagger$ \\                   
Mean & 0.765 & 0.340 & 0.210 & 0.340 & 0.193 & 0.020 & 0.016 & 0.017 \\
Std & 0.009 & 0.024 & 0.014 & 0.024 & 0.012 & 0.003 & 0.002 & 0.002 \\
\textbf{FairBatch-Eqopp} $\dagger$ \\
Mean & 0.755 & 0.351 & 0.212 & 0.351 & 0.191 & 0.022 & 0.018 & 0.019 \\
Std & 0.009 & 0.015 & 0.008 & 0.015 & 0.008 & 0.002 & 0.003 & 0.003 \\
$\text{\textbf{JTT}}^+$ & \\
Mean & 0.767 & \underline{0.306} & \underline{0.194} & \underline{0.306} & 0.191 & \underline{0.016} & \underline{0.015} & \underline{0.016} \\
Std & 0.007 & 0.038 & 0.017 & 0.038 & 0.007 & 0.005 & 0.001 & 0.001 \\
\midrule 
\textbf{DFL-demog} $\dagger$ &  \\
Mean & 0.762 & \textbf{0.241} & \textbf{0.124} & \textbf{0.241} & 0.233 & \textbf{0.000} & \textbf{0.003} & \textbf{0.003} \\
Std & 0.006 & 0.006 & 0.004 & 0.006 & 0.009 & 0.000 & 0.000 & 0.000 \\
\textbf{BLIND} &  \\
Mean & 0.761 & 0.403 & 0.240 & 0.403 & \underline{0.190} & 0.030 & 0.020 & 0.021 \\
Std & 0.008 & 0.034 & 0.020 & 0.034 & 0.005 & 0.006 & 0.005 & 0.004 \\
\midrule 
\textbf{Control} &  \\
Mean & 0.780 & 0.443 & 0.253 & 0.443 & 0.175 & 0.038 & 0.026 & 0.025 \\
Std & 0.007 & 0.012 & 0.010 & 0.012 & 0.005 & 0.002 & 0.003 & 0.004 \\
\toprule
\end{tabular}
}
\caption{ 
Full results on the sentiment classification task for BERT. For each fairness metric, the best result is \textbf{bolded}, and the best result achieved without demographics in the training data is \underline{underlined}.}
\label{tbl:full_results_moji_bert}
\end{table*}

\begin{table*}[ht]
\adjustbox{max width=\textwidth}{%
\begin{tabular}{lllllllll}
\toprule
{} & \multicolumn{8}{c}{DeBERTa} \\
\cmidrule(lr){2-9}
{} & Acc $\uparrow$ & TPR (s) &  $\text{TPR}_{RMS}$ $\downarrow$ & $\text{FPR}_s$ $\downarrow$ & $\text{Prec}_s$ $\downarrow$ & Indep $\downarrow$ & Sep $\downarrow$ & Suff $\downarrow$ \\
\midrule
\textbf{Finetuned} \\
Mean & 0.775 & 0.482 & 0.270 & 0.482 & 0.198 & 0.047 & 0.033 & 0.032 \\
Std & 0.005 & 0.061 & 0.031 & 0.061 & 0.031 & 0.013 & 0.006 & 0.006 \\
\textbf{INLP} $\dagger$ &  \\
Mean & 0.633 & \textbf{0.149} & \textbf{0.086} & \textbf{0.149} & \textbf{0.100} & \textbf{0.010} & 0.010 & \textbf{0.015} \\
Std & 0.161 & 0.205 & 0.118 & 0.205 & 0.084 & 0.015 & 0.013 & 0.012 \\
\textbf{RLACE} $\dagger$ & \\
Mean & 0.748 & 0.297 & 0.168 & 0.297 & 0.225 & 0.016 & 0.008 & 0.011 \\
Std & 0.017 & 0.098 & 0.059 & 0.098 & 0.031 & 0.015 & 0.011 & 0.008 \\
\textbf{FairBatch-DP} $\dagger$ \\
Mean & 0.763 & 0.317 & 0.203 & 0.317 & 0.207 & 0.016 & 0.016 & 0.017 \\
Std & 0.012 & 0.011 & 0.007 & 0.011 & 0.005 & 0.001 & 0.004 & 0.004 \\
\textbf{FairBatch-Eqodds} $\dagger$ \\                   
Mean & 0.768 & 0.324 & 0.210 & 0.324 & 0.205 & 0.018 & 0.014 & 0.015 \\
Std & 0.010 & 0.043 & 0.024 & 0.043 & 0.009 & 0.005 & 0.002 & 0.002 \\
\textbf{FairBatch-Eqopp} $\dagger$ \\
Mean & 0.758 & 0.323 & 0.203 & 0.323 & 0.204 & 0.017 & 0.018 & 0.019 \\
Std & 0.011 & 0.031 & 0.017 & 0.031 & 0.002 & 0.004 & 0.003 & 0.003 \\
$\text{\textbf{JTT}}^+$ & \\
Mean & 0.770 & \underline{0.276} & \underline{0.185} & \underline{0.276} & 0.202 & \underline{0.012} & \underline{0.013} & \underline{0.014} \\
Std & 0.011 & 0.042 & 0.018 & 0.042 & 0.009 & 0.005 & 0.003 & 0.003 \\
\midrule 
\textbf{DFL-demog} $\dagger$ &  \\
Mean & 0.766 & 0.265 & 0.177 & 0.265 & 0.211 & \textbf{0.009} & 0.006 & 0.008 \\
Std & 0.017 & 0.008 & 0.010 & 0.008 & 0.015 & 0.004 & 0.004 & 0.003 \\
\textbf{BLIND} &  \\
Mean & 0.782 & 0.396 & 0.243 & 0.396 & \underline{0.200} & 0.028 & 0.022 & 0.022 \\
Std & 0.006 & 0.017 & 0.010 & 0.017 & 0.012 & 0.003 & 0.001 & 0.001 \\
\midrule 
\textbf{Control} &  \\
Mean & 0.783 & 0.415 & 0.243 & 0.415 & 0.179 & 0.033 & 0.026 & 0.025 \\
Std & 0.009 & 0.048 & 0.021 & 0.048 & 0.012 & 0.009 & 0.006 & 0.006 \\
\bottomrule
\toprule
\end{tabular}
}
\caption{ 
Full results on the sentiment classification task for DeBERTa. For each fairness metric, the best result is \textbf{bolded}, and the best result achieved without demographics in the training data is \underline{underlined}.}
\label{tbl:full_results_moji_deberta}
\end{table*}
\begin{table*}[ht]
\adjustbox{max width=\textwidth}{%

\begin{tabular}{llllllllllll}
\bottomrule
\toprule
{} & \multicolumn{11}{c}{BERT} \\
\cmidrule(lr){2-12}
{} & Acc $\uparrow$ & TPR (s) &  $\text{TPR}_{RMS}$ $\downarrow$ & $\text{TPR}_p$ $\downarrow$ & $\text{FPR}_s$ $\downarrow$ & $\text{FPR}_p$ $\downarrow$ & $\text{Prec}_s$ $\downarrow$ & $\text{Prec}_p$ $\downarrow$ & Indep $\downarrow$ & Sep $\downarrow$ & Suff $\downarrow$ \\
\midrule
\textbf{Finetuned} \\ 
Mean & 0.86 & 2.46 & 0.14 & 0.81 & 0.07 & 0.57 & 3.80 & -0.90 & 0.01 & 0.34 & 1.56 \\
Std & 0.00 & 0.27 & 0.01 & 0.03 & 0.01 & 0.04 & 0.17 & 0.03 & 0.00 & 0.09 & 0.15 \\
\textbf{INLP} $\dagger$ &  \\
Mean & 0.85 & 2.48 & 0.13 & 0.78 & 0.07 & 0.58 & 3.26 & -0.86 & 0.01 & 0.33 & 1.22 \\
Std & 0.01 & 0.43 & 0.02 & 0.07 & 0.02 & 0.04 & 0.37 & 0.02 & 0.00 & 0.07 & 0.07 \\
\textbf{RLACE} $\dagger$ & \\
Mean & 0.87 & 2.44 & 0.13 & 0.81 & 0.07 & 0.63 & 3.65 & -0.90 & 0.01 & 0.35 & 1.41 \\
Std & 0.00 & 0.33 & 0.01 & 0.05 & 0.00 & 0.02 & 0.29 & 0.02 & 0.00 & 0.04 & 0.16 \\
$\text{\textbf{JTT}}^+$ & \\
Mean & 0.85 & \underline{2.45} & 0.14 & 0.76 & \underline{0.07} & 0.60 & 3.43 & -0.88 & 0.01 & 0.35 & 1.42 \\
Std & 0.00 & 0.31 & 0.02 & 0.02 & 0.01 & 0.02 & 0.31 & 0.02 & 0.00 & 0.07 & 0.15 \\
\textbf{Scrubbing} $\dagger$ & \\
Mean & 0.86 & 2.15 & 0.12 & 0.70 & 0.06 & 0.49 & 2.77 & -0.84 & 0.01 & 0.27 & 0.90 \\
Std & 0.00 & 0.34 & 0.01 & 0.07 & 0.00 & 0.07 & 0.31 & 0.04 & 0.00 & 0.03 & 0.14 \\
\midrule 
\textbf{DFL-demog} $\dagger$ &  \\
Mean & 0.83 & 2.09 & 0.12 & 0.60 & \textbf{0.06} & 0.41 & 2.38 & -0.78 & \textbf{0.00} & 0.25 & 0.71 \\
Std & 0.01 & 0.22 & 0.01 & 0.13 & 0.01 & 0.11 & 0.28 & 0.02 & 0.00 & 0.07 & 0.13 \\
\textbf{DFL-demog + Scrubbing} $\dagger$\\
Mean & 0.83 & \textbf{1.88} & \textbf{0.11} & \textbf{0.57} & \textbf{0.06} & \textbf{0.38} & \textbf{2.18} & \textbf{-0.72} & \textbf{0.00} & \textbf{0.20} & \textbf{0.58} \\
Std & 0.00 & 0.19 & 0.01 & 0.12 & 0.01 & 0.06 & 0.43 & 0.06 & 0.00 & 0.04 & 0.18 \\
\textbf{BLIND} &  \\
Mean & 0.83 & 2.47 & 0.14 & \underline{0.69} & 0.08 & \underline{0.47} & \underline{3.09} & \underline{-0.77} & 0.01 & \underline{0.32} & \underline{1.10} \\
Std & 0.00 & 0.20 & 0.01 & 0.07 & 0.02 & 0.11 & 0.85 & 0.09 & 0.00 & 0.08 & 0.47 \\
\textbf{BLIND + Scrubbing} &  \\
Mean & 0.84 & 2.39 & 0.13 & 0.66 & 0.07 & 0.51 & 2.89 & -0.79 & 0.01 & 0.33 & 1.07 \\
Std & 0.00 & 0.20 & 0.01 & 0.09 & 0.01 & 0.05 & 0.48 & 0.04 & 0.00 & 0.05 & 0.28 \\
\midrule 
\textbf{Control} &  \\
Mean & 0.86 & 2.64 & 0.14 & 0.80 & 0.08 & 0.60 & 3.90 & -0.88 & 0.01 & 0.37 & 1.60 \\
Std & 0.00 & 0.19 & 0.01 & 0.01 & 0.01 & 0.03 & 0.64 & 0.04 & 0.00 & 0.05 & 0.34 \\
\toprule
\end{tabular}
}
\caption{ 
Full results on the occupation classification task for BERT. For each fairness metric, the best result is \textbf{bolded}, and the best result achieved without demographics in the training data is \underline{underlined}.}
\label{tbl:full_results_bios_bert}
\end{table*}

\begin{table*}[ht]
\adjustbox{max width=\textwidth}{%
\begin{tabular}{llllllllllll}
{} & \multicolumn{11}{c}{DeBERTa} \\
\cmidrule(lr){2-12}
{} & Acc $\uparrow$ & TPR (s) &  $\text{TPR}_{RMS}$ $\downarrow$ & $\text{TPR}_p$ $\downarrow$ & $\text{FPR}_s$ $\downarrow$ & $\text{FPR}_p$ $\downarrow$ & $\text{Prec}_s$ $\downarrow$ & $\text{Prec}_p$ $\downarrow$ & Indep $\downarrow$ & Sep $\downarrow$ & Suff $\downarrow$ \\
\midrule
\textbf{Finetuned} \\
Mean & 0.86 & 2.49 & 0.13 & 0.82 & 0.08 & 0.55 & 3.73 & -0.87 & 0.01 & 0.35 & 1.60 \\
Std & 0.00 & 0.46 & 0.03 & 0.02 & 0.01 & 0.06 & 0.22 & 0.03 & 0.00 & 0.07 & 0.13 \\
\textbf{INLP} $\dagger$ &  \\
Mean & 0.85 & 2.11 & 0.12 & 0.73 & 0.07 & 0.56 & 3.14 & -0.84 & 0.01 & 0.31 & 1.05 \\
Std & 0.01 & 0.18 & 0.01 & 0.04 & 0.01 & 0.02 & 0.27 & 0.04 & 0.00 & 0.03 & 0.11 \\
\textbf{RLACE} $\dagger$ & \\
Mean & 0.87 & 2.34 & 0.13 & 0.81 & 0.07 & 0.63 & 3.65 & -0.89 & 0.01 & 0.33 & 1.36 \\
Std & 0.00 & 0.28 & 0.01 & 0.01 & 0.00 & 0.02 & 0.28 & 0.03 & 0.00 & 0.04 & 0.11 \\
\textbf{Scrubbing} $\dagger$ & \\
Mean & 0.86 & 1.95 & 0.11 & 0.73 & 0.06 & 0.56 & 2.73 & -0.86 & 0.01 & 0.27 & 0.79 \\
Std & 0.01 & 0.18 & 0.01 & 0.05 & 0.01 & 0.02 & 0.21 & 0.03 & 0.00 & 0.04 & 0.09 \\
$\text{\textbf{JTT}}^+$ & \\
Mean & 0.84 & 2.51 & 0.14 & 0.77 & 0.08 & 0.58 & 3.53 & -0.86 & 0.01 & 0.35 & 1.40 \\
Std & 0.00 & 0.27 & 0.01 & 0.03 & 0.00 & 0.05 & 0.31 & 0.03 & 0.00 & 0.05 & 0.27 \\
\midrule 
\textbf{DFL-demog} $\dagger$ &  \\
Mean & 0.83 & 1.93 & \textbf{0.11} & 0.62 & 0.06 & 0.43 & 2.69 & -0.78 & \textbf{0.00} & 0.26 & 0.79 \\
Std & 0.00 & 0.15 & 0.01 & 0.06 & 0.00 & 0.07 & 0.42 & 0.04 & 0.00 & 0.03 & 0.25 \\
\textbf{DFL-demog + Scrubbing} $\dagger$\\
Mean & 0.84 & \textbf{1.90} & \textbf{0.11} & \textbf{0.58} & \textbf{0.05} & \textbf{0.36} & \textbf{2.13} & \textbf{-0.70} & \textbf{0.00} & \textbf{0.24} & \textbf{0.59} \\
Std & 0.01 & 0.18 & 0.01 & 0.08 & 0.01 & 0.09 & 0.30 & 0.07 & 0.00 & 0.04 & 0.15 \\
\textbf{BLIND} &  \\
Mean & 0.84 & \underline{2.16} & \underline{0.12} & \underline{0.64} & \underline{0.08} & \underline{0.51} & \underline{2.98} & \underline{-0.81} & \underline{0.01} & \underline{0.31} & \underline{0.91} \\
Std & 0.00 & 0.20 & 0.01 & 0.08 & 0.01 & 0.05 & 0.30 & 0.04 & 0.00 & 0.03 & 0.21 \\
\textbf{BLIND + Scrubbing} &  \\
Mean & 0.84 & 2.03 & 0.12 & 0.65 & 0.06 & 0.39 & 2.48 & -0.81 & 0.00 & 0.27 & 0.71 \\
Std & 0.00 & 0.22 & 0.01 & 0.12 & 0.00 & 0.05 & 0.32 & 0.07 & 0.00 & 0.05 & 0.14 \\
\midrule 
\textbf{Control} &  \\
Mean & 0.87 & 2.49 & 0.13 & 0.81 & 0.07 & 0.49 & 3.51 & -0.89 & 0.01 & 0.34 & 1.38 \\
Std & 0.00 & 0.59 & 0.02 & 0.10 & 0.01 & 0.06 & 0.38 & 0.03 & 0.00 & 0.07 & 0.38 \\
\bottomrule
\toprule
\end{tabular}
}
\caption{ 
    Full results on the occupation classification task for DeBERTa. For each fairness metric, the best result is \textbf{bolded}, and the best result achieved without demographics in the training data is \underline{underlined}.}
\label{tbl:full_results_bios_deberta}
\end{table*}

\end{document}